\documentclass[review]{elsarticle}

\usepackage{hyperref}
\usepackage{float}
\usepackage{graphicx}
 \usepackage{amsmath,amssymb}
 \usepackage{color}
 \usepackage{url}
 \usepackage{soul,color}
 \newcommand{\UrbanObject}{u}
 \newcommand{\Class}{l}
 \newcommand{\Picture}{x}
 \newcommand{\Feature}{f}
 \newcommand{\Aggregated}{g}
 
 \newcommand{\pred}[1]{\widehat{#1}}
 \newcommand{\V}[1]{\mathbf{#1}}
 
\newcommand{\norm}[1]{\left\lVert#1\right\rVert}
\newcommand{\shiva}[2]{\textcolor{black}{#2}}
\newcommand{\shivangi}[2]{\textcolor{black}{#2}}

\journal{Remote Sensing of Environment}

\begin{document}

\begin{frontmatter}

\title{Understanding urban landuse from the above and ground perspectives: a deep learning, multimodal solution}

\author[wur]{Shivangi Srivastava}
\author[unicamp]{John E. Vargas-Mu\~{n}oz}
\author[wur]{Devis Tuia}
\cortext[grant]{First Author: SS, shivangi.srivastava@wur.nl.}

\address[wur]{Laboratory of Geo-information Science and Remote Sensing, Wageningen University \& Research, the Netherlands}
\address[unicamp]{Laboratory of Image Data Science, Institute of Computing, University of Campinas, Campinas, Brazil}

\begin{abstract}
\textbf{This is the pre-acceptance version, to read the final version published in the journal Remote Sensing of Environment, please go to: \url{https://doi.org/10.1016/j.rse.2019.04.014}}
Landuse characterization is important for urban planning. It is traditionally performed with field surveys or manual photo interpretation, two practices that are time-consuming and labor-intensive. Therefore, we aim to automate landuse mapping at the urban-object level with a deep learning approach based on data from multiple sources (or \emph{modalities}). We consider two image modalities: overhead imagery from Google Maps and ensembles of ground-based pictures (side-views) per urban-object from Google Street View (GSV). These modalities bring complementary visual information pertaining to the urban-objects. We propose an end-to-end trainable model, which uses OpenStreetMap annotations as labels. The model can accommodate a variable number of GSV pictures for the ground-based branch and can also function in the absence of ground pictures at prediction time. We test the effectiveness of our model over the area of \^Ile-de-France, France\shiva{}{, and test its generalization abilities on a set of urban-objects from the city of Nantes, France}. Our proposed multimodal Convolutional Neural Network achieves considerably higher accuracies than methods that use a single image modality, making it suitable for automatic landuse map updates. Additionally, our approach could be easily scaled to multiple cities, because it is based on data sources available for many cities worldwide.
\end{abstract}


\end{frontmatter}


\section{Introduction and Related Work}

According to the UN report ``The World’s Cities in 2016"\footnote{\url{http://www.un.org/en/development/desa/population/publications/pdf/urbanization/the_worlds_cities_in_2016_data_booklet.pdf}}, the population living in urban areas will rise from 4 billions in 2016 to a projected 5 billions in 2030. Therefore, it becomes important to gather information about how land is being utilized in urban areas. This information provides insights to city planners, helping them managing current urban infrastructure as well as planning for future cities. In this paper, landuse is defined as the utility of a particular area for humans: for example, an area could be used as a school, a park, a museum or a hospital. The mapping of various landuses is traditionally done through field surveys, which are often time consuming, expensive and labor intensive to carry out. This makes it impractical to frequently update these maps. Therefore, it is imperative to design models capable of automating the generation of landuse maps using data-driven approaches.

In the last decade, great advances have been observed for the automation of landcover maps using remote sensing imagery~\cite{homer2015completion,postadjiana2017dnnclasfctnhrimagery,inglada2017operational} 
and current large scale efforts extend this logic to multiple cities worldwide~\cite{Tau12,Dem18}. Landcover \shiva{implies}{mapping considers the characterization of} various materials visible on the Earth's surface, for example, crops, orchards, forests, water bodies, roads or buildings.  \shiva{}{Earlier solutions to the problem classified each pixel based solely on its spectral signature~\cite{Riggan2009}, since this information is correlated with the underlying material.In cases where the spectral information would not be sufficient to discriminate between landcover classes, contextual and texture information~\cite{Myint2001robust} were integrated, usually by analyzing a fixed size window around each pixel. Later, unsupervised segmentation methods were widely used to partition the image and perform object-based classification, allowing to extract more discriminative features and also contextual information from neighbor regions~\cite{Blaschke_2014, Ma2017object}. More recently, Convolutional Neural Networks (CNN) have attained more accurate classification results \cite{Zhu17}. CNNs learn in a supervised way, a hierarchy of filters to extract high-level features, using both spectral and spatial information. They have been used to perform classification in a patch-based way~\cite{DFCA,Sharma2017,Tui17f} and also to classify all the pixels of the input image in one forward pass~\cite{volpi2017densesemanticlabeling,Audebert2016semantic}.}

\begin{figure}
\centering
\includegraphics[width=0.98\linewidth, height=0.24\textheight]{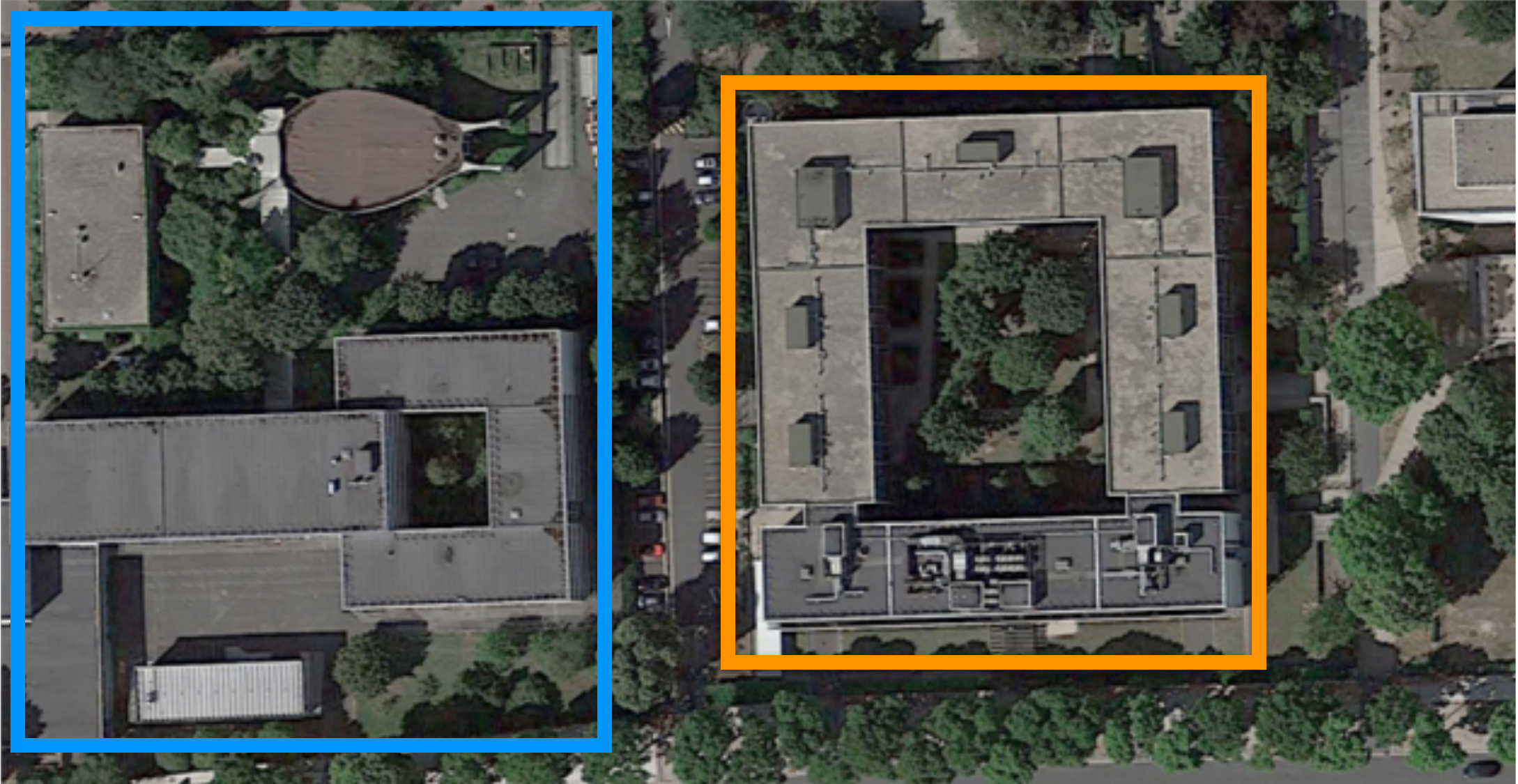}
\caption{\label{fig:overheadEducationalGovernmentParis} This view from the top shows an educational institute building on the left (blue) and a government building on the right (orange). The two landuse classes are difficult to discern using only remotely sensed imagery. Source: Imagery from Google Maps of an area in the city of Paris.}
\end{figure}

\shiva{}{F}ollowing a similar approach \shiva{}{based on overhead images only} to generate accurate large scale landuse\footnote{\shivangi{}{We define landuse as the way in which a delimited geographical space is utilized by humans. For example, this might be a hospital, a school, a museum, a park, etc.}} maps is not an easy task, because the spectral signature of materials alone is not sufficient for discerning different landuse types. The problem is two-fold: 1) most of the times, a landuse class is made of a combination of different landcover types. For example, a university could have in its premises buildings, trees, grass, water bodies and roads. 2) The same landcover types are observed across multiple landuse classes. For example, when seen from above, similar building architectures could be a government office or a school
(see Figure~\ref{fig:overheadEducationalGovernmentParis}). 
Therefore, generating an accurate landuse map at the urban-object\footnote{We define an urban-object as a spatial construct in an urban space with a clear physical boundary of its own, which could be a closed construct (like shop, office), semi-open construct (like stadium), or an open space (e.g. a natural forest or man-made park).} level from overhead imagery alone is a challenging task. Still, some works have been done in this direction, typically following a patch-based classification scheme~\cite{castelluccio2015landuse,hu2015transferring,Her12,Vol14,Bec15} \shiva{}{or hybrid approaches that involves patch- and object-based analysis~\cite{Zhang_2018}}. A typical pattern in these studies is the search for more representative feature spaces to describe landuse, for instance using textures and context~\cite{pacifici2009neural,tuia2015multiclass} or higher order information~\cite{Vol18,Mar17}. The assumption is that, when seen from the top, different landuse types show different structural characteristics. Some recent works also explored the use of data from other sources, such as road networks or OpenStreetMap\footnote{\url{https://www.openstreetmap.org/}} (OSM) vector data~\cite{Yok17}. The assumption in these cases is that the remotely sensed information alone is insufficient in describing landuse, and that the incorporation of complementary, meaningful data sources is beneficial.

In parallel, researchers have also approached the landuse mapping problem from the ground perspective, typically by using other data sources such as ground based pictures from online repositories (e.g. Flickr, Instagram, Geograph)~\cite{leung2012exploring,zhu2015landuseCNNgbpics,tracewski2017repurposingdnnphoto,zhu2018landusegoogleplacesapi}. The ground-based viewpoint of these pictures provides crucial information on the function of urban-objects conventionally hidden from the view above, such as school entrances. However, the pictures from these repositories also have shortcomings: 1) they are often not accurately geo-referenced; 2) they sometimes depict highly personalized content (mostly touristic viewpoints, selfies or zoomed objects) rather than visual information about the urban-object; 3) they tend to cover the city unevenly (most pictures are geo-located in touristic areas). These problems make such pictures databases less suitable for our purpose, i.e., reliable landuse mapping of a city. Nonetheless, thanks to the availability of services like Google Street View\footnote{\url{ https://developers.google.com/maps/documentation/streetview/}} (GSV), it is nowadays possible to obtain ground-based pictures for many urban-objects with objective content, which are accurately geo-located and are densely available across many cities worldwide. These GSV pictures are also updated regularly and it is possible to access historical data. GSV pictures have proven to be beneficial for complex tasks such as urban trees detection~\cite{wegner2016catalogingpublicobjects} or detection of urban fabric changes~\cite{naik2017cvuncoversphysicalchange}. For a review of recent papers dealing with aerial to ground fusion tasks, please refer to~\cite{Lef17}.

GSV is also being increasingly used in landuse classification~\cite{srivastava2018_ijgis,yair2015clascatngsvstorefronts,kang2018buildinginstance,workman2017farandnearrs}. Authors in~\cite{yair2015clascatngsvstorefronts} used a deep Convolutional Neural Network (CNN) to perform store front classification in 13 business categories from single GSV pictures. Authors in~\cite{kang2018buildinginstance} classify the landuse of urban-objects into 8 classes by using GSV pictures and labels from OSM. The model predicts one label for each picture in the set of GSV pictures corresponding to one urban-object. The final predicted label corresponds to the class with the maximum average classification score. This last strategy might be suboptimal for our case: since the model learns landuse of an urban-object from pictures considered independently, thus it will force images with similar typical objects (e.g. pictures with trees) to be classified into different landuse classes. This makes training unnecessarily difficult and  leaves the final decision to the majority vote, which can succeed only under a strong assumption: that each urban-object of a class will be imaged mostly with pictures containing objects that are both typical and unique for that specific class. Instead, we argue that each landuse category is made of different objects present in a set of images:  in our previous work~\cite{srivastava2018_ijgis}, we proposed a model that learns class representations from ensembles of GSV pictures. In this paper, we extend it to a multi-modal strategy, leveraging the complementarity of aerial and terrestrial views.

Landuse mapping using both terrestrial pictures and remote sensing data is a new and emerging field: to the best of our knowledge, the only paper dealing with it explicitly is~\cite{workman2017farandnearrs} over New York City, by means of landuse labels provided by the New York City Department of City Planning. Using footprints and labels from authoritative sources makes the method less scalable to cities where such building footprints (and their landuse labels) could be either sparse, of insufficient quality or may have strongly variable landuse definitions across cities. Another important difference is that their proposed model performed per-pixel classification. The feature representation of each pixel was obtained using a fixed number $N_{loc}$ of nearby locations, where street level panoramas were available. For each of these $N_{loc}$ locations, GSV pictures looking in the four cardinal directions were used. A drawback of this approach is that  pictures taken in such way provide features that may depict objects unrelated to the landuse observed at the pixel level.

In this paper, we  learn a multimodal model leveraging visual information from both aerial and ground views to predict landuse
at an urban-object level. Looking at the growing success of deep learning algorithms in remote sensing~\cite{Zhu17}, we propose a  model that combines visual information of overhead imagery and ground-based pictures associated with the urban-objects and trains end-to-end. The urban-object footprints and the ground truth labels are collected from OSM. We study the effectiveness of the proposed model on a case study in the region of {\^I}le-de-France (France). Our proposed model outperforms architectures based on unimodal data. This shows the importance and complementarity of both the data sources. For most landuse categories, the proposed multimodal model obtains accuracies above 70\%. 

Since GSV images are not always available or can be of insufficient quality (for instance by positioning errors or occlusions), we also propose a module able to process urban-objects for which the GSV images are missing: by using a joint three-view \emph{embedding} space that projects into a common representation, the deep features obtained for two modalities (a set of GSV pictures and the overhead imagery imaging the same urban-object)  and landuse categories data for each urban-object. This embedding space is useful, since it allows to perform cross-modality retrieval: by looking for nearest neighbors, the system is able to retrieve from the training set the most likely GSV feature vector for the urban-object and use it for prediction.

By combining standard deep learning building blocks in a new efficient way and using solely widely available data, our model can be easily deployed and also be transferred to new urban environments, where OSM annotations are available. The main contributions of the work are:
\begin{itemize}
\item[-] The development of a deep learning system based on widely available data to describe landuse classes at the urban-object level;
\item[-] The design of a system that accepts a variable number of street-level images to describe appearance from multiple points of view;
\item[-] The addition of an embedding module making the system robust to the lack of ground-based pictures for an urban-object at test time. In that case, an alternative ensemble of plausible GSV pictures from the training set is retrieved and used together with the overhead imagery to predict the landuse class accurately.
\end{itemize}

The paper is organized as follows: In Section~\ref{sec:model} we present the proposed model in detail. Section~\ref{sec:data} brings forward how the dataset was created for the region of {\^I}le-de-France. Section~\ref{sec:experiments} shows the experimental setup while results are discussed in Section~\ref{sec:results}. Section~\ref{sec:conclusions} concludes the paper.

\section{Methods}\label{sec:model}
In this paper, we define landuse classification as the task of predicting a class label $\Class_\UrbanObject \in [1, ..., K]$ of a given urban-object $\UrbanObject$, where $K$ is the number of landuse classes. In our case, each urban-object is defined by a polygon footprint obtained from OSM (see Section~\ref{sec:data}), along with its label (also from OSM). In order to predict the category of the urban-object $\UrbanObject$, we have a collection of $N_u$ ground-based pictures $\{\V{\Picture}_{\UrbanObject}^i\}_{i=1}^{N_u}$ and one overhead image $\V{o}_u$ of this urban-object. The procedure to collect this dataset is discussed in Section~\ref{sec:data}.

Our proposed Convolutional Neural Network model is composed of two streams: the `Overhead Imagery Stream' and the `Ground-based Pictures Stream' (see Figure~\ref{fig:siameseAndOverheadModel}), that extracts discriminative features from overhead imagery and ground-based pictures, respectively. The features learned for the two streams are then combined to perform the prediction of the final landuse category. \shivangi{}{Note that we are not aiming at performing semantic segmentation at the pixel level, but our objective is rather to predict the landuse category of the urban-objects, which are vectorial objects in OpenStreetMap. }In Sections~\ref{ssec:overheadCNN} and~\ref{ssec:siam}, we describe the two CNN models that are used with either modality (these unimodal CNN models are also our baselines for comparison). In Section~\ref{ssec:multimodal}, we show how our proposed model combines the two streams to perform landuse classification. In ~\ref{ssec:cca} we discuss how to use
a  projective method based on canonical correlations to cope with situation where the GSV modality is not available at test time.

\subsection{CNN Architecture for Overhead Imagery}\label{ssec:overheadCNN}

\begin{figure}
\centering
\includegraphics[width=\linewidth, height=\textheight,keepaspectratio]{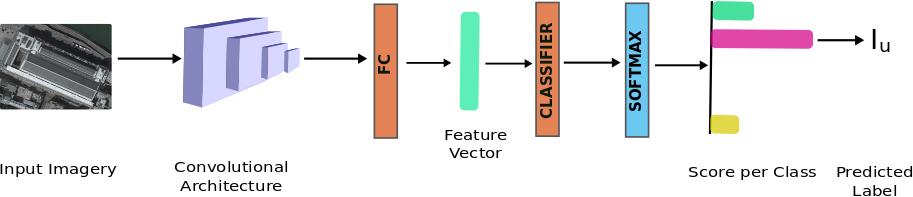}
\caption{Overhead image classification architecture.}
\label{fig:overheadModel}
\end{figure}

This first baseline accepts remote sensing imagery and is thus related to traditional patch-based remote sensing image classification methods (e.g.~\cite{penatti2015cvprw}). For every OSM footprint, we use an overhead image crop that covers it completely. Figure~\ref{fig:overheadModel} depicts our corresponding CNN architecture. The overhead imagery is used as an input for a sequence of convolutional blocks {(violet part in Figure~\ref{fig:overheadModel}}, with each block encompassing a convolution operation, followed by spatial pooling and a non-linear activation function ({Rectified Linear Unit; ReLU}) that outputs an activation map. Then, a fully connected layer converts the activation map into a high-dimensional feature vector (in green). Another fully connected layer is then applied that projects the feature vector into class scores; these are eventually normalized to $[0,1]$ by means of a softmax operation. The category with maximum score is considered as the final predicted class. Several works~\cite{castelluccio2015landuse,hu2015transferring} have shown good landuse classification performance by fine-tuning CNN models that were trained in large data sets for object recognition (i.e., ImageNet~\cite{russakovsky2015imagenet}). Similarly, we used the popular VGG16 architecture~\cite{simonyan2014cnnlargescaleimagereco} pretrained on ImageNet as a base trunk to extract features (in violet in Figure~\ref{fig:overheadModel}). 

\subsection{Siamese-like Architecture for Ground Based Pictures}\label{ssec:siam}

Urban-objects are generally surrounded by roads, which allows us to associate multiple GSV pictures to them. This means that for such an OSM footprint, we get discriminative and complementary representations thanks to GSV pictures capturing its object from different points of view. In our previous work~\cite{srivastava2018_ijgis} we exploited this observation and proposed the Variable Input Siamese Convolutional Neural Network (VIS-CNN). This model learns a single feature representation of an arbitrary number of GSV pictures for a given urban-object in an end-to-end manner.
\begin{figure}
\centering
\includegraphics[width=\linewidth, height=\textheight,keepaspectratio]{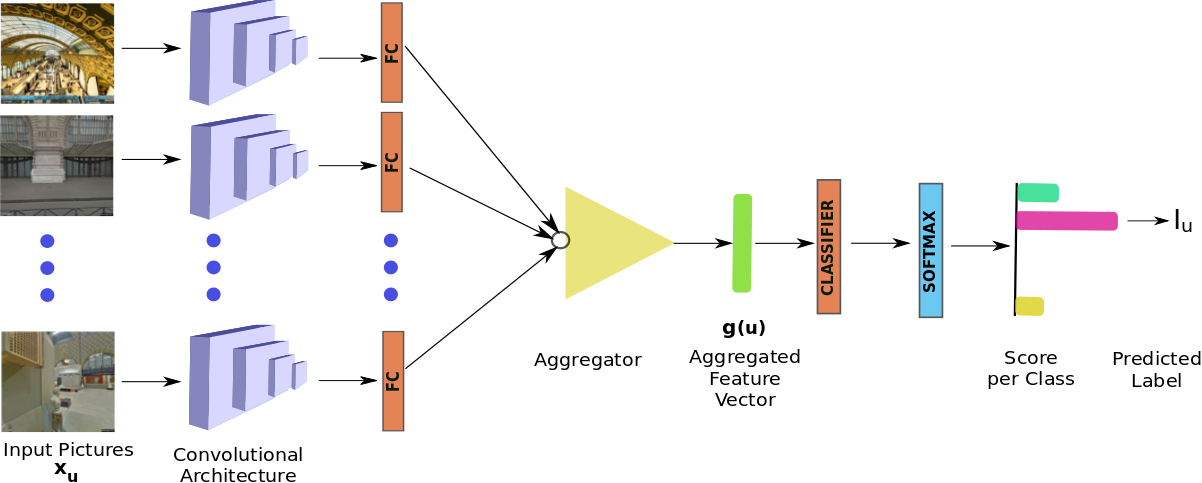}
\caption{Variable Input Siamese Convolutional Neural Network (VIS-CNN) for ground based pictures.}
\label{fig:siameseModel}
\end{figure}
Figure~\ref{fig:siameseModel} depicts the VIS-CNN model for landuse classification using ground-based pictures. First, the convolutional blocks and the fully connected layers extract the feature vectors for each image. Note that the same CNN model (VGG16~\cite{simonyan2014cnnlargescaleimagereco}, pre-trained on the ImageNet dataset) is used for each image to extract these features. Afterwards, the $N_u$ feature vectors $\V{\Feature(\Picture}_\UrbanObject^i)$, one per each picture $i$, pertaining to urban-object $u$, are aggregated to obtain a single feature descriptor of the urban-object $u$. In~\cite{srivastava2018_ijgis} we compared aggregation strategies based on average and max pooling:
\begin{align}
\Aggregated(\UrbanObject)_\textrm{max}^j &= \max\limits_{i} \Feature(\V{\Picture}_\UrbanObject^i)^j\label{eq:max}\,,\\
\Aggregated(\UrbanObject)_\textrm{avg}^j &= \frac{1}{N_\UrbanObject}\sum\limits_{i=1}^{N_\UrbanObject}\Feature(\V{\Picture}_\UrbanObject^i)^j\,,\label{eq:avg}
\end{align}
where $\Feature(\cdot)^j$ is the $j^{th}$ element of the vector $\V{\Feature(\cdot)}$. The \emph{max} operator performs input selection picking the most important representation, among all the pictures, per element in the feature vector. The \emph{avg} aggregator assigns importance to the most repeated attributes among all the pictures associated with the urban-object. Experimentally, we had observed that the \emph{avg} aggregator peforms better than the \emph{max}~\cite{srivastava2018_ijgis}, thus we will use \emph{avg} aggregator in the experiments below. Interestingly, this is also in line with very recent results obtained in the field of image deblurring from image sequences~\cite{Ait18}, where the authors proposed a very similar architecture as ours to cope with the problem of variability of the length of the sequence.

Finally, the computed aggregated vector $\V{\Aggregated}(\UrbanObject)$ is used as input of the last fully connected layer (classifier), that outputs the classification scores for each category to obtain the final prediction.  

\subsection{Multimodal CNN Architecture}\label{ssec:multimodal}
 
The two models described in the previous sections have very similar bottlenecks, both corresponding to a \emph{d}-dimensional fully connected layer. In this section, we take advantage of this similarity in order to perform late representation fusion.

\begin{figure}
\centering
\includegraphics[width=\linewidth, height=\textheight,keepaspectratio]{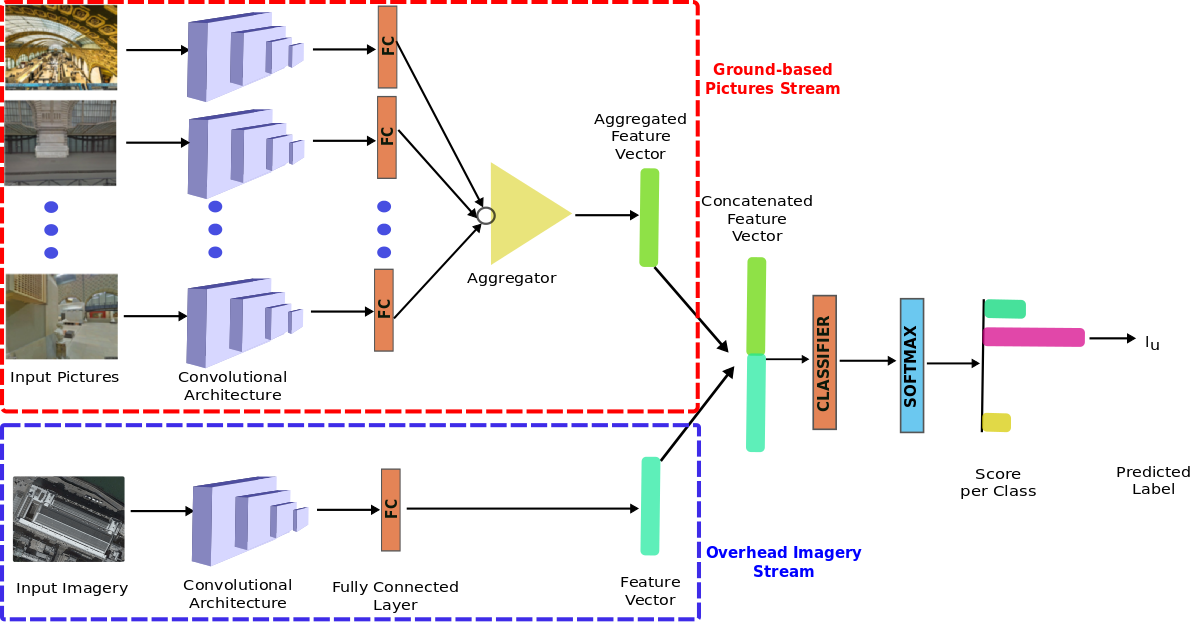}
\caption{Proposed Multimodal Model with two streams (highlighted with dashed red and {blue} lines). Our model extracts features from both modalities, namely ground-based pictures (red) and overhead imagery (blue). The extracted features from both streams are concatenated to finally predict the landuse category.}
\label{fig:siameseAndOverheadModel}
\end{figure}

Figure~\ref{fig:siameseAndOverheadModel} depicts the proposed CNN model for multimodal landuse classification. For every urban-object $u$ we use its corresponding set of $N_u$ ground-based pictures $\{\V{\Picture}_{\UrbanObject}^i\}_{i=1}^{N_u}$ (used as inputs for the model described in Section~\ref{ssec:siam}), as well as its corresponding overhead imagery $\V{o}_u$ (used as input of the model described in Section~\ref{ssec:overheadCNN}). In both cases, we stop at the level of feature extraction, i.e. we remove the classifiers in the architectures illustrated in Figures~\ref{fig:overheadModel} and Figure~\ref{fig:siameseModel} and only keep the convolutional blocks for feature extraction. Then, the image features are combined by a fully connected layer that outputs a score for each landuse category. After that, a softmax layer is applied to obtain normalized classification scores as for the previous models. 

In order to learn the parameters of the CNN model, we use the cross-entropy loss function:
\begin{multline}
L  = \frac{1}{N} \sum_{u=1}^{N} \Biggl[-\sigma(\pred{\Class_u} = \Class_{u}|\V{\Picture}_u^1,\ldots,\V{\Picture}_u^{N_u}, \V{o}_u) +
\log\biggl( \sum_{k=1}^{K} \exp(\sigma(\pred{\Class_u} = k| \V{\Picture}_u^1,\ldots,\V{\Picture}_u^{N_u}, \V{o}_u)) \biggr) \Biggr] \,,\label{eq:loss}
\end{multline}
where $\sigma(\pred{\Class_u} = k| \V{\Picture}_u^1,\ldots,\V{\Picture}_u^{N_u}, \V{o}_u)$ is the softmax score given by the model for the urban-object $u$ and class $k$.

\subsection{Missing Modality Retrieval with Three-View CCA}\label{ssec:cca}

\begin{figure}[!t]
\centering
\includegraphics[width=1.0\linewidth,height=0.64\textheight]{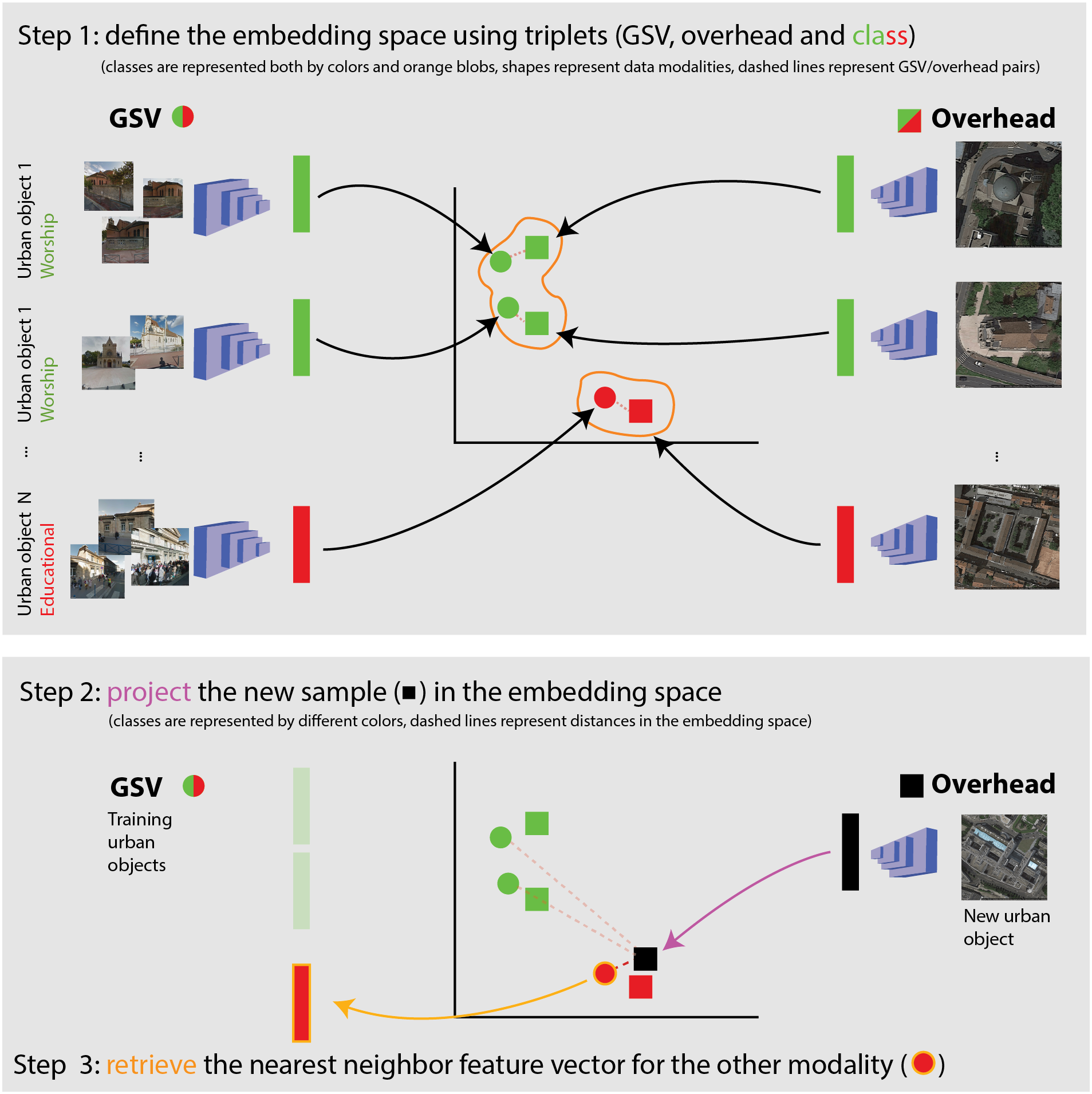}
\caption{Intuition behind the CCA embedding for retrieving missing GSV features.}\label{fig:embed}
\end{figure}

In this section, we present a solution to cope with urban-objects, for which no street level picture is available at test time. We limit analyses to this case, as a situation with missing overhead imagery is less likely to happen. However, the approach is general and could as well be applied to such a scenario. We propose to compensate for the missing modality by retrieving the closest train GSV feature vector for the queried test overhead imagery feature vector. The GSV pictures for the retrieved closest GSV feature and the overhead imagery of the urban-object are used in situ as an input to the proposed multimodal model (see Section~\ref{ssec:multimodal}). The missing GSV modality retrieval task can be broadly divided into three steps ({}{also} illustrated in Figure~\ref{fig:embed}):
\begin{enumerate}
\item Define the projection matrices for the joint embedding space by using the features extracted by the two CNN models (see Sections~\ref{ssec:overheadCNN}~and~\ref{ssec:siam}) on the training set.
\item Use these matrices to project the overhead CNN features for the test sample in the same embedding space.
\item Given the \emph{overhead} projected features, find the nearest projected \emph{GSV} feature neighbor from the training set. Which in turn, gives the nearest urban-object from the train set that we consider a proxy of what the urban-object would have looked like in GSV pictures. Once found, use the GSV pictures of this nearest neighbor urban-object in the multimodal model.
\end{enumerate}

To define the joint embedding space, we exploit the fact that we have paired views of ensemble of GSV pictures and one overhead imagery for each urban-object in the training set, along with its landuse class. Under this assumption, we can define a space where two views (features from set of GSV pictures and top-view imagery) for an urban-object are projected close to each other and far from those of urban-objects belonging to different classes. This is possible because we are using class information that allows samples of the same class to be projected closer than samples coming from other land use classes (a typical assumption in this type of projective methods~\citep{Tui13d,Tui16d}). To this end, we use a projective technique based on Canonical Correlation Analysis (CCA~\citep{Nielsen1998,Vol14b}). 

We have three datasets: $\hat{X}_1$ and $\hat{X}_2$ are the features issued from the two views (GSV and overhead imagery), while $X_3$ corresponds to the class labels. Each row of $\hat{X}_1$, $\hat{X}_2$ and $X_3$ represents a feature vector coming from three different modalities, but representing the same object. Originally, the dimensions of the three dataset are $(N \times 4096)$ for $\hat{X}_1$, $(N \times 4096)$ for $\hat{X}_2$, and $(N \times 16)$ for $X_3$ (the sixteen classes labels are encoded as a sixteen dimensional one-hot vector, with $1$ for the correct class and $0$ otherwise). To decrease the size of the matrices involved in the eigenvalue decomposition problem involved in CCA, a Principal Component Analysis (PCA) is applied to matrices $\hat{X}_1$ and $\hat{X}_2$ separately. This is a common practice in nonlinear dimensionality reduction, since embedding high-dimensional spaces is very difficult because of the curse of dimensionality and the noise in high dimensional data~\citep{Lee07}. In the following, we refer to the matrices obtained after PCA reduction as $X_1$ with size $(N \times d_1)$ and $X_2$ with size $(N \times d_2)$, where $d_1, d_2 < 4096$.  

CCA finds projection matrices $W_i$ (one per view, $i=1,2,3$) that project the features $X_i$ from the view-specific spaces into a low-dimensional common embedding space, in which the distances between different views for the same data item are minimized (Equation~\eqref{eq:CCAobjFunctnWithoutKernel}). The objective function for this problem can be written as :

\begin{equation}
\begin{aligned}
& \underset{W_1 , W_2, W_3}{\text{min}}
\quad \sum_{i,j=1}^{3} {\norm{X_iW_i - X_jW_j}}_F^2, \\
& \text{subject to}
\quad W_i^T\Sigma_{ii}W_i = \mathit{I} , w_{ik}^T\Sigma_{ij}w_{jl} = 0 \\
& i,j = 1,2,3 ,\; i \neq j \quad k,l = 1, \ldots ,d, \; k \neq l \\
\end{aligned}
\label{eq:CCAobjFunctnWithoutKernel}
\end{equation}

where $\Sigma_{ii}$ is the covariance matrix of $X_i$ and $w_{ik}$ is the $k^{th}$ column of $W_i$. This problem can be solved as the following generalized eigenvalues problems as in Equation~\eqref{eq:EigenValueDecomp} (see~\cite{gong2014cca} for details):

\begin{eqnarray}
\begin{pmatrix}
C_{11} & C_{12} & C_{13}\\
C_{21} & C_{22} & C_{23}\\
C_{31} & C_{32} & C_{33}
\end{pmatrix}
\begin{pmatrix}
w_{1} \\
w_{2} \\
w_{3}
\end{pmatrix} =
\begin{pmatrix}
C_{11} & 0 & 0\\
0 & C_{22} & 0 \\
0 & 0 & C_{33}
\end{pmatrix} 
\begin{pmatrix}
w_{1} \\
w_{2} \\
w_{3}
\end{pmatrix},
\label{eq:EigenValueDecomp}
\end{eqnarray}

where $C_{ij} = X_i^TX_j$ is the covariance matrix between the $i^{th}$ and $j^{th}$ views and $w_i$ is a column of $W_i$. The size of this problem (corresponding to the maximal size of the embedding space) is $(d_1 + d_2 + d_3) \times (d_1 + d_2 + d_3)$ where $d_i$ is the dimensionality of the respective input data spaces (in our case, $4096$ for the CNN trained on GSV, $4096$ for the CNN trained on the overhead images and $16$ for the classes term). Also, a regularization parameter $\eta$ = $10^{-4}$ is added to the diagonal of the covariance matrix $C_{ij}$ to better condition the problem.

Once the projection matrices $W_i$ are learned (using the training set) by solving Equation~\eqref{eq:EigenValueDecomp}, we can use them to project new, unseen test data into the latent space and assess their relative position with respect to samples from the training data (Step 2 in Figure~\ref{fig:embed}). In our case, we want to project CNN features from the overhead view of the test urban-object in the joint embedding space, in order to retrieve the closest GSV feature vector.  Usually, only the first few dimensions of the projected space are relevant for expressing correlations across views~\citep{Vol14b}. Hence it is a common practice to use only the top eigenvectors to define the projection matrices. In order to do this, we keep the top $d_{emb} << d_1+d_2+d_3$ eigenvectors as projection matrices $W_1$, $W_2$ and $W_3$. After this selection, the projection matrices have dimensionality: $W_1 \in R^{d_1 \times d_{emb}}$, $W_2 \in R^{d_2 \times d_{emb}}$ and, $W_3 \in R^{d_3 \times d_{emb}}$.

After projection, we can assess similarities between the projected vectors ($X_2^*W_2$) of overhead data in test set ($X_2^*$) and those coming from GSV in training set ($X_1W_1$). To do so, we use the similarity function used in~\cite{gong2014cca} as it leads to greater retrieval accuracy compared to that using Euclidean distance:
\begin{equation}
\begin{aligned}
& sim(X_1, X_2^*) = \frac{(X_1W_1D_1)(X_2^*W_2D_2)^T}{\norm{(X_1W_1D_1)}_2 \norm{(X_2^*W_2D_2)}_2} \\
\end{aligned}
\label{eq:cos_similarity}
\end{equation}
where $W_i$ is the projection matrix and  $D_i$ is a diagonal matrix containing $d_{emb}$ eigenvalues, with each entry raised to the power $p$ ~\cite{gong2014cca, chapelle2002clusterkernels}. Now, for any projected overhead imagery feature in the test set, we can query the closest projected GSV feature in the training set that minimizes Equation~\eqref{eq:cos_similarity}. The GSV pictures from the urban-object (corresponding to the resulting nearest GSV feature) together with the overhead imagery are used as input to the proposed multi-modal model (Figure~\ref{fig:siameseAndOverheadModel}). This way, we obtain the final label prediction as presented in Section~\ref{ssec:multimodal}.

\section{Dataset}
\label{sec:data}
In order to evaluate our proposed method, we collected data from OSM, Google Maps and GSV in the region of {\^I}le-de-France, France. For this study we considered the metropolitan area of Paris and the nearby suburbs including Versailles, Orsay, Orly, Aulnay-sous-Bois, Le Bourget, Sarcelles, Chatou and Nanterre. For the supervised training stage of our multimodal CNN, we created an annotated dataset, which is made of an ensemble of side-view pictures and one overhead image view per urban-object with their corresponding landuse ground truth. The data collection procedure is detailed in the following subsections. \shiva{}{Additionally, and in order to evaluate the generalization ability of the model trained with {\^I}le-de-France data, we have also gathered data and evaluated our method over the city of Nantes.}

\subsection{Annotations from OSM}

\begin{figure}
\begin{center}
\includegraphics[width=1\textwidth]{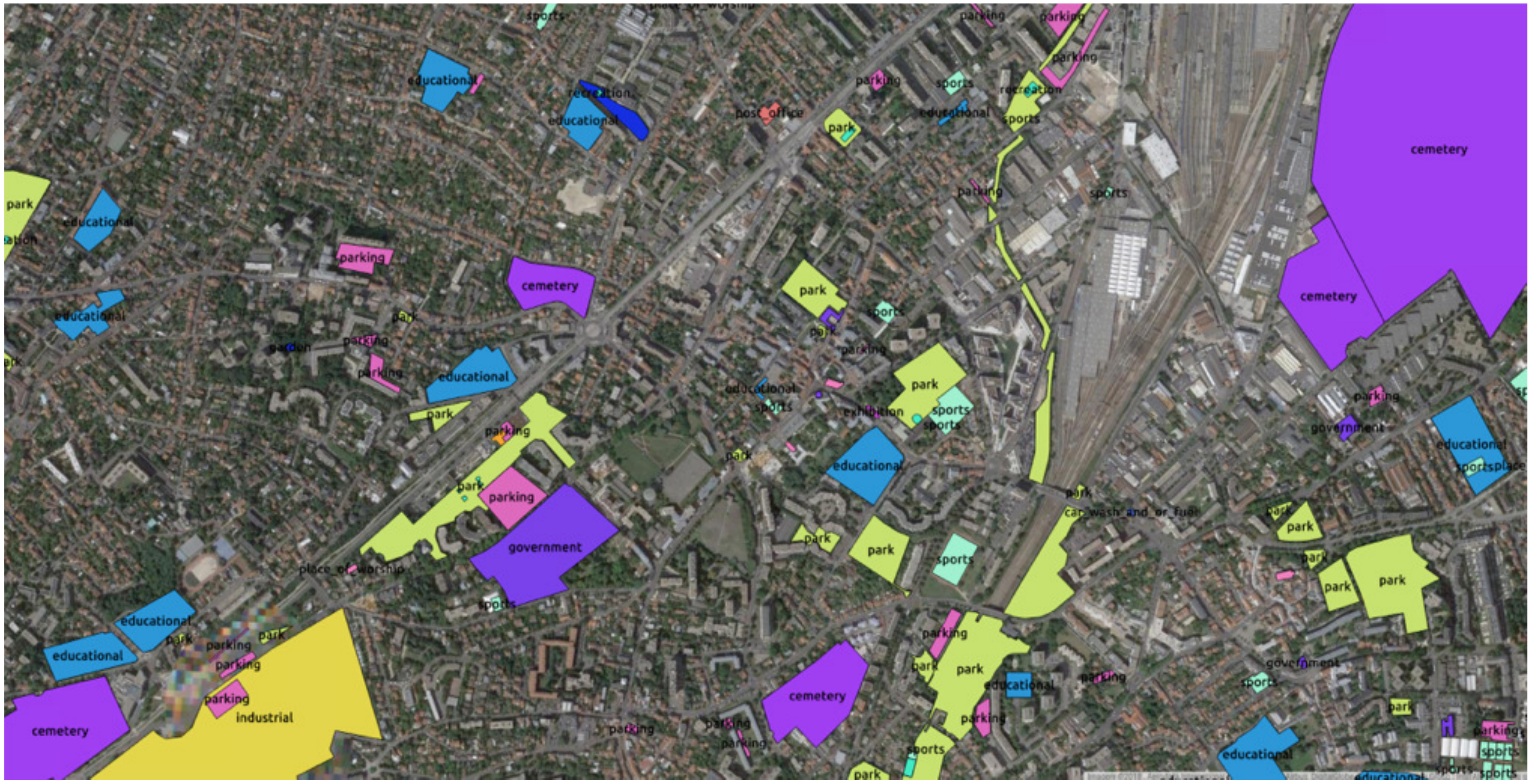}
\caption{Some examples of footprints of urban-objects as obtained from OSM with their corresponding landuse labels.}
\label{fig:dataset1}
\end{center}
\end{figure}

\shivangi{}{We use OSM to obtain a collection of urban-objects with associated landuse categories. We group OSM landuse categories into 16 classes based on the similarity of their ``usage" (For example, ``lyc\'ee" and ``\'ecole" are merged into a single class, ``educational". Synagogues and churches are merged into the class ``religious"). Rarely appearing landuse classes like ``crematorium" or ``observatory" are not considered due to the limited amount of OSM footprints or of  the corresponding GSV pictures. The selected 16 landuse classes are:} ``educational'', ``hospital'', ``religious'', ``shop'', ``cemetery'', ``forest'', ``park'', ``heritage'', ``sports'', ``government'', ``post office'', ``parking, ``fuel'', ``marina'', ``hotel'', ``industrial''.
We collected the spatial footprints and landuse labels of the selected OSM polygons. Labels were processed for consistency and disambiguation~\cite{srivastava2018_ijgis}. \shiva{Our dataset consists of $5941$ urban-objects from the 16 classes of which a subset is depicted in Figure~\ref{fig:dataset1}.}{Two datasets are created, the first containing $5941$ urban-objects from the region of {\^I}le-de-France. A subset of this data is depicted in Figure~\ref{fig:dataset1}. The second datasets contains $1835$ urban-objects from the city of Nantes. Both datasets contain the same landuse classes, with the exception of the class ``Marina" in the city of Nantes, that was omitted due to the lack of urban-objects available (only one urban-object was retrieved from OSM for Nantes).}

\subsection{Ground-based Pictures and Overhead Imagery}
\begin{figure}
\begin{center}
\includegraphics[width=1.0\linewidth,height=0.35\textheight]{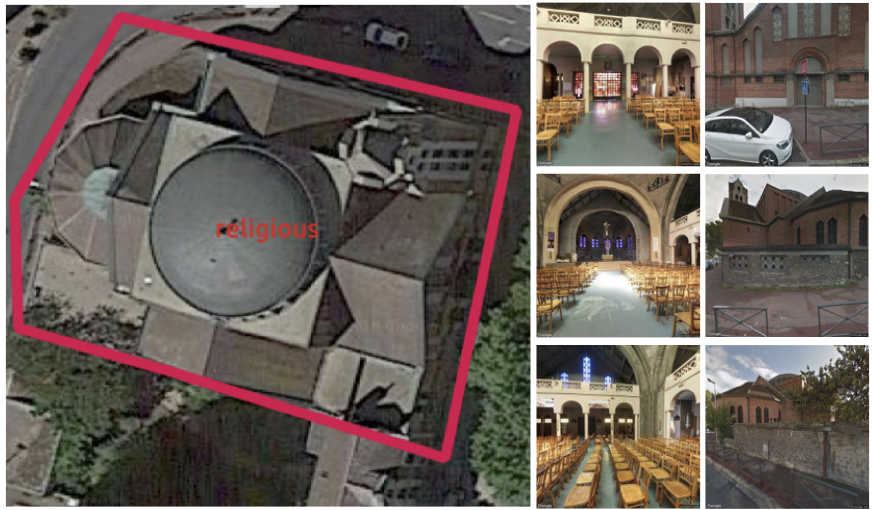}
\caption{The left panel shows the overhead imagery of the urban-object delineated by a red box, with the corresponding landuse class from OSM. The six images on the right show some of the GSV pictures for the same urban-object.}
\label{fig:dataset2}
\end{center}
\end{figure}

To obtain the ground-based pictures corresponding to each urban-object, we used the Google Street View API. We downloaded a set of pictures from various viewpoints (Figure~\ref{fig:dataset2}) in the following way: to collect the images oriented towards the urban-object, we selected the roads nearest to that urban-object and downloaded pictures (of size $640\times 640$ pixels) looking at the fa\c{c}ade of the urban-object from different viewpoints and at a distance of maximum 12 meters from the object itself. Additionally, pictures located within the urban-object (which are often uploaded by users) were also retrieved using the same API. In this last situation, and when applicable, we downloaded pictures for inside locations in the four cardinal directions. \shiva{For the $5941$ urban-objects present in the OSM footprints dataset, we downloaded a total of $44957$ GSV pictures.}{For the $5941$ urban-objects present in the OSM footprints dataset of {\^I}le-de-France, we downloaded a total of $44957$ GSV pictures, while for the $1835$ urban-objects corresponding to Nantes we downloaded $9908$ GSV pictures.}

Regarding the aerial images, we used the Google Maps Static API to obtain the top-view image of each urban-object, ensuring that the downloaded imagery covered the entire footprint. The original downloaded images have size of $1280\times 1280$ pixels, with ground pixel resolution depending on the width of the urban-object footprint. We downsampled the overhead images to  $240\times 240$ pixels to be used in the CNN model. \shiva{The number of overhead images corresponds to the number of footprints, i.e. $5941$.}{The number of overhead images corresponds to the number of footprints, i.e. $5941$ for {\^I}le-de-France and $1835$ for Nantes.}

\section{Experimental Setup}\label{sec:experiments}
\subsection{Joint CNN Training}
To extract features from each image, we used the VGG16 model~\cite{simonyan2014cnnlargescaleimagereco}, both for the multimodal CNN and the baselines. For all models, the hyperparameters were kept fixed and the models were trained end-to-end with the following settings: the number of urban-objects processed in each training iteration was $4$, while the initial learning rate was set to $0.001$. Further, the learning rate was divided by a factor of 10 after every 10 epochs. The training was pursued for 50 epochs with Stochastic Gradient Descent (SGD) with momentum~\cite{krizhevsky2012imagenet} as an optimizer. For data augmentation, we used the following strategies:
\begin{itemize}
\item We resized the GSV pictures to $256 \times 256$ pixels, followed by random crops of size $224 \times 224$ pixels. The cropped image underwent random horizontal flipping and was normalized using the mean and the standard deviation values from the ImageNet dataset. 
\item The overhead images were downscaled to $240 \times 240$ pixels and randomly flipped in both vertical and horizontal directions, to strengthen invariance in the model. 
\end{itemize}
The dataset was split into five different train and test sets. For each split, we \shivangi{kept}{randomly selected} 80\% of the urban-objects per landuse class for training and the remaining was set aside for testing. \shivangi{}{Note that the train and test sets are mutually exclusive.} We calculate overall accuracy (OA) and average of accuracy per class (AA) over the test set in each split. The averaged OA and AA over 5 splits per model is presented in Table~\ref{table:accuracy_scores}. All the experiments were run on a server running Linux and featuring a GeForce GTX 1080 Ti GPU. We used the PyTorch CNN library\footnote{\url{http://pytorch.org/}} for the computations. The time to train the multimodal model for 50 epochs was between $15-16$ hours, while the Siamese model took between $11-12$ hours and the overhead model was trained in $3-4$ hours.

\subsection{Missing modality retrieval}
After studying the ability of the system to predict landuse, we examined the possibility of using the CCA-based retrieval algorithm presented in Section~\ref{ssec:multimodal} to process urban-objects for which GSV data are not available. 
As detailed in the methodology section, we used the training data to define the embedding space. The features were extracted by using the VIS-CNN model (Section~\ref{ssec:siam}) for GSV pictures and VGG16 for the overhead imagery (Section~\ref{ssec:overheadCNN}). The feature vectors were normalized by dividing each one by its $L2$ norm. The CCA system has three hyperparameters, which we fixed empirically:
\begin{itemize}
\item[-] \emph{$\%pca$} is the percentage of total feature dimension kept after applying PCA. The resulting dimensions of data matrices $X_1$ and $X_2$  are $N \times d_1 $ and $ N \times d_2$ respectively, where $d_1, d_2 = 410$ (10\% of $4`096$). For the label matrix $X_3$, we keep $d_3 = 16$. The final dimension of eigenvalue decomposition (equation~\ref{eq:EigenValueDecomp}) decreases from $8`208$ to $836$.

\item[-] $\%d_{emb} = \frac{ d_{emb}}{ d_1 + d_2 + d_3}$ is the percentage of eigenvectors kept to compute the projection matrices and corresponds to the final dimension of the embedding space. It was chosen empirically as $\%d_{emb} = 0.2$.

\item[-] \emph{$p$} is the power of the eigenvalues matrices $D_i$ in Equation~\eqref{eq:cos_similarity}. It was chosen empirically as $p = 6$.

\end{itemize}
We will also present  a study of the sensitivity of the free parameters in Section~\ref{ssec:cca}.

\section{Results and Discussion}\label{sec:results}
\subsection{Joint CNN Training}
\begin{figure}
\centering
\includegraphics[width=1\textwidth]{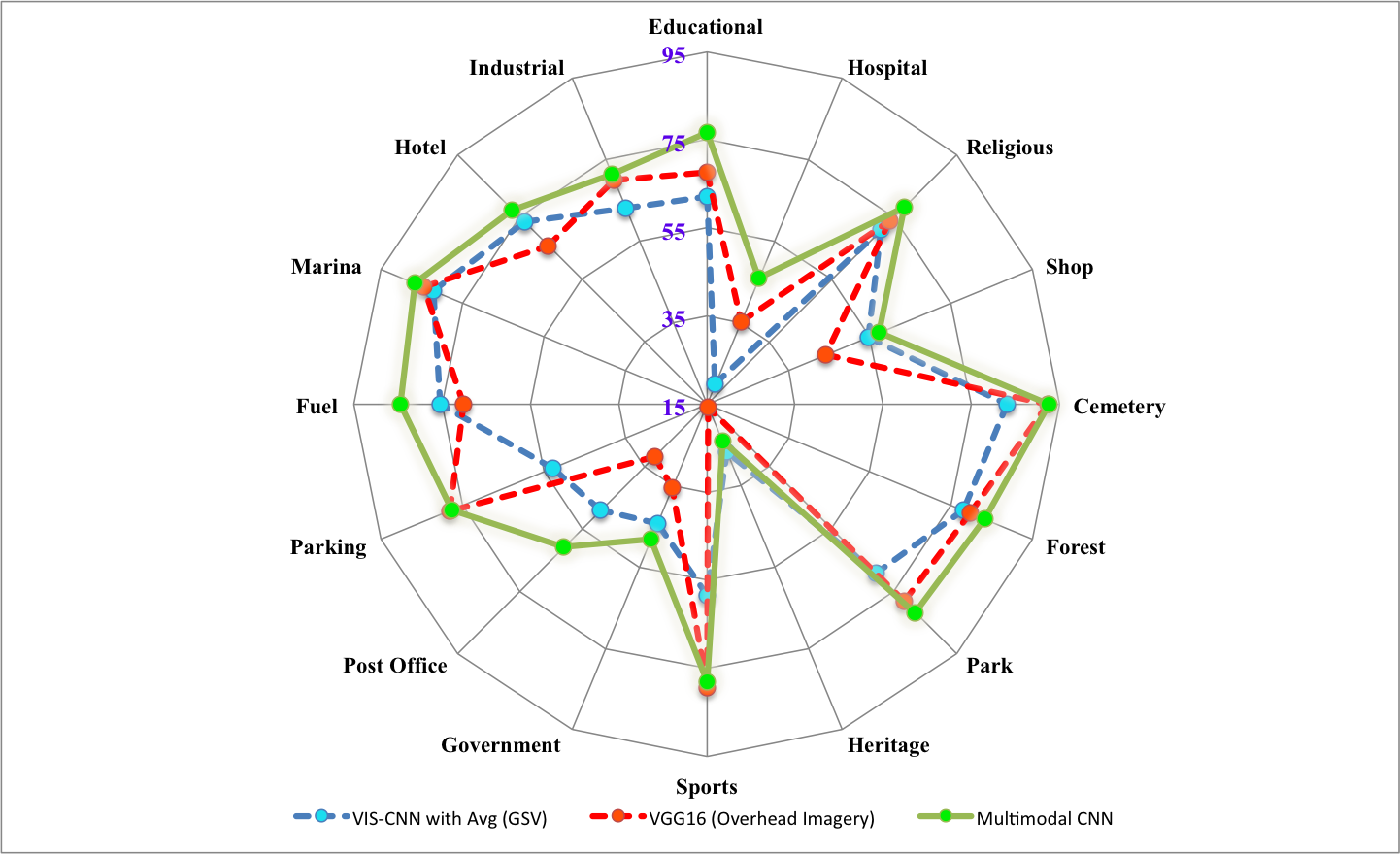}
\caption{Class-specific accuracy scores for the three models compared in Table~\ref{table:accuracy_scores}.}
\label{fig:radar_results}
\end{figure}

The class accuracies are shown in Figure~\ref{fig:radar_results}; averaged OA and AA values are given in Table~\ref{table:accuracy_scores}. By comparing our multimodal model against the unimodal variants, we observe an increase of around 6\% for OA and more than 7\% for AA against the VGG16{-based model} trained on overhead imagery, while a sharp increase of more than 10\% for both OA and AA is observed when comparing with VIS-CNN trained on GSV pictures. \shiva{}{Additionally, we evaluated our proposed Multimodal CNN and VIS-CNN using different base CNN models. Specifically, AlexNet~\cite{krizhevsky2012imagenet} that was used in~\cite{Huang_2018} to perform landuse mapping with mutltispectral remote sensing images and ResNet50~\cite{He_2016} that was used in~\cite{Tong_2018} to do large-scale land cover classification of satellite imagery. The results of these methods are presented in Table~\ref{table:accuracy_scores_other_cnns}. Similar gains in performance are observed for the Mutilmodal CNN with respect to the unimodal models.}

\setlength{\tabcolsep}{3.2pt}
\begin{table}[t]
\begin{center}
\caption{Accuracy scores for our proposed Multimodal CNN model and two unimodal CNN models (OH: overhead imagery, GSV: Google Street View ground based images, \shiva{}{rGSV: GSV feature vectors retrieved through the CCA algorithm}. OA: overall accuracy; AA: average accuracy) \shiva{}{for   {the \^I}le-de-France dataset}}
\label{table:accuracy_scores}
\begin{tabular}{l|c|c|cc}
\hline\noalign{\smallskip}
&\multicolumn{2}{c|}{Data source(s)}&\multicolumn{2}{c}{Metric} \\
Model Name & Train & Test  &  OA & AA\\
\noalign{\smallskip}
\hline
\noalign{\smallskip}
VGG16~\cite{simonyan2014cnnlargescaleimagereco}&OH & OH & 67.48 $\pm$0.57 & 62.67$\pm$1.39 \\
VIS-CNN with Avg~\cite{srivastava2018_ijgis} & GSV & GSV & 62.52 $\pm$1.12 & 60.24 $\pm$1.71 \\
Multimodal CNN & OH, GSV & OH, GSV & \textbf{73.44} $\pm$0.96 & \textbf{70.30} $\pm$2.59 \\\hline 
\multicolumn{1}{c}{ }\\
Multimodal CNN + CCA & OH, GSV & OH, \shiva{retrieved GSV}{rGSV} &{71.78} $\pm$1.02 & {65.65} $\pm$1.71\\
\hline
\end{tabular}
\end{center}
\end{table}
\setlength{\tabcolsep}{1.4pt}

Looking at the per-class predictions (Figure~\ref{fig:radar_results}), we can observe that our proposed multimodal model outperforms the baselines for almost all of the classes. Landuse classes like \emph{educational}, \emph{hospital}, \emph{post-office} and \emph{fuel} benefit from a jump of more than 9\%, while classes like \emph{religious} and \emph{hotel} see an increase of more than 4\% in their accuracies. 

\begin{figure}
\centering
\includegraphics[width=1.0\textwidth]
{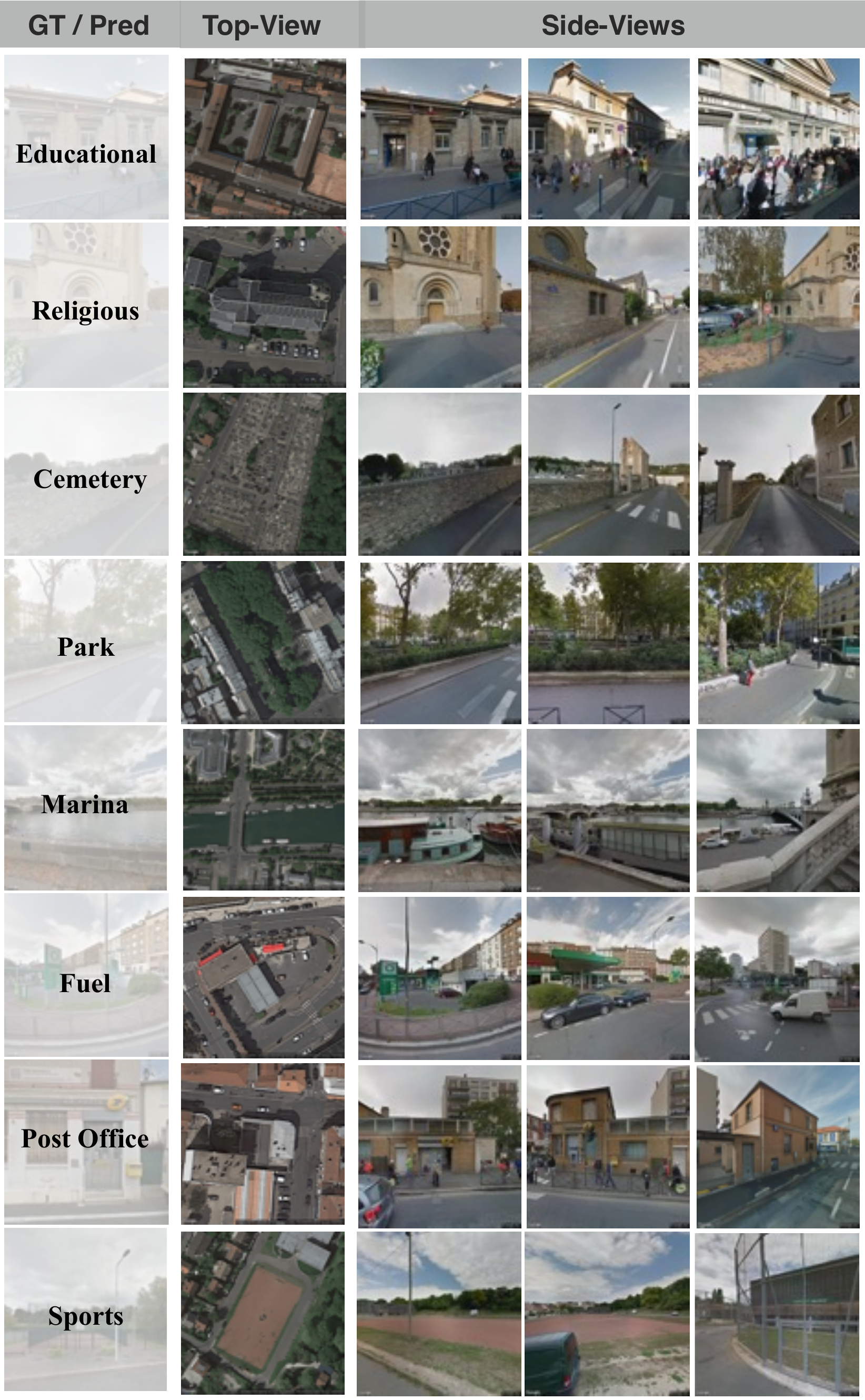}
\caption{\label{fig:positive_results} Correct classifications by the proposed multimodal CNN model (first column), with examples of both the overhead imagery (second column) and GSV pictures (third to fifth columns) involved. Each row represents a single urban-object.}
\end{figure}

Some of the correct predictions of our model can be seen in Figure~\ref{fig:positive_results}. For each example, we discuss briefly the complementary visual cues that are used by the multimodal model to predict the landuse category.
For the class \emph{educational} (with accuracy 77\%), objects like playgrounds within the school campus are visible in overhead imagery. This information complements the one brought by the side-views, including flags, a big entrance, the presence of metal fencing and broad pedestrian walks, or the presence of children (first row, Figure~\ref{fig:positive_results}). If we analyze the overhead imageries pertaining to religious places (accuracy 78\%), we notice stylized roofs with absence of pipes, chimneys, exhausts, and the like. This adds complementary information to the big arched doors, rose windows and stained glasses coming from the ground pictures (Figure~\ref{fig:positive_results}, second row). 
The third row in Figure~\ref{fig:positive_results} shows the overhead imagery and set of GSV pictures for a correctly predicted sample for class \emph{cemetery}, which has a very high accuracy (92\%). We can observe several visual cues in the overhead imagery, like the specific grid pattern of the grave stones, separated by wide alleys. This has been complemented by the ground views, which contain visible long continuous walls typical for cemeteries. Finally, in the case of the post office (accuracy 61\%), the overhead imagery shows yellow delivery vans in the parking close by. This adds to characteristic visual objects  that are usually present in the ground pictures, like the yellow ``la-poste" signboard (seventh row, Figure~\ref{fig:positive_results}).

Classes like \emph{government} and \emph{shop}, despite having training sets of 400 and 267 objects respectively, have comparatively lower accuracy scores (see Figure~\ref{fig:radar_results}) for all the models. In the case of the multimodal model, the accuracies are still around 48\% and 57\%, respectively. Surprisingly, for the class \emph{fuel}, though the number of training samples is only 122, its accuracy score is much higher (84.5\%). We attribute the good result for the class fuel to the distinctive visual information from both ground and top views (see sixth row, Figure~\ref{fig:positive_results}), which allows the CNN to perform well, even in the absence of a large dataset. On the contrary, classes like \emph{heritage sites} and \emph{sports} show a very small decrease in their accuracy scores compared to the VIS-CNN (for GSV pictures) and VGG16 (overhead imagery), respectively (see Figure~\ref{fig:radar_results}). In the case of \emph{heritage sites}, the overhead imagery does not carry discriminative information from the top view (
{as evident through} the poor accuracy of 15.8\% for the overhead model), which degrades the quality of the multimodal result as well.

\begin{figure}[t]
\centering
\includegraphics[width=1.0\textwidth]{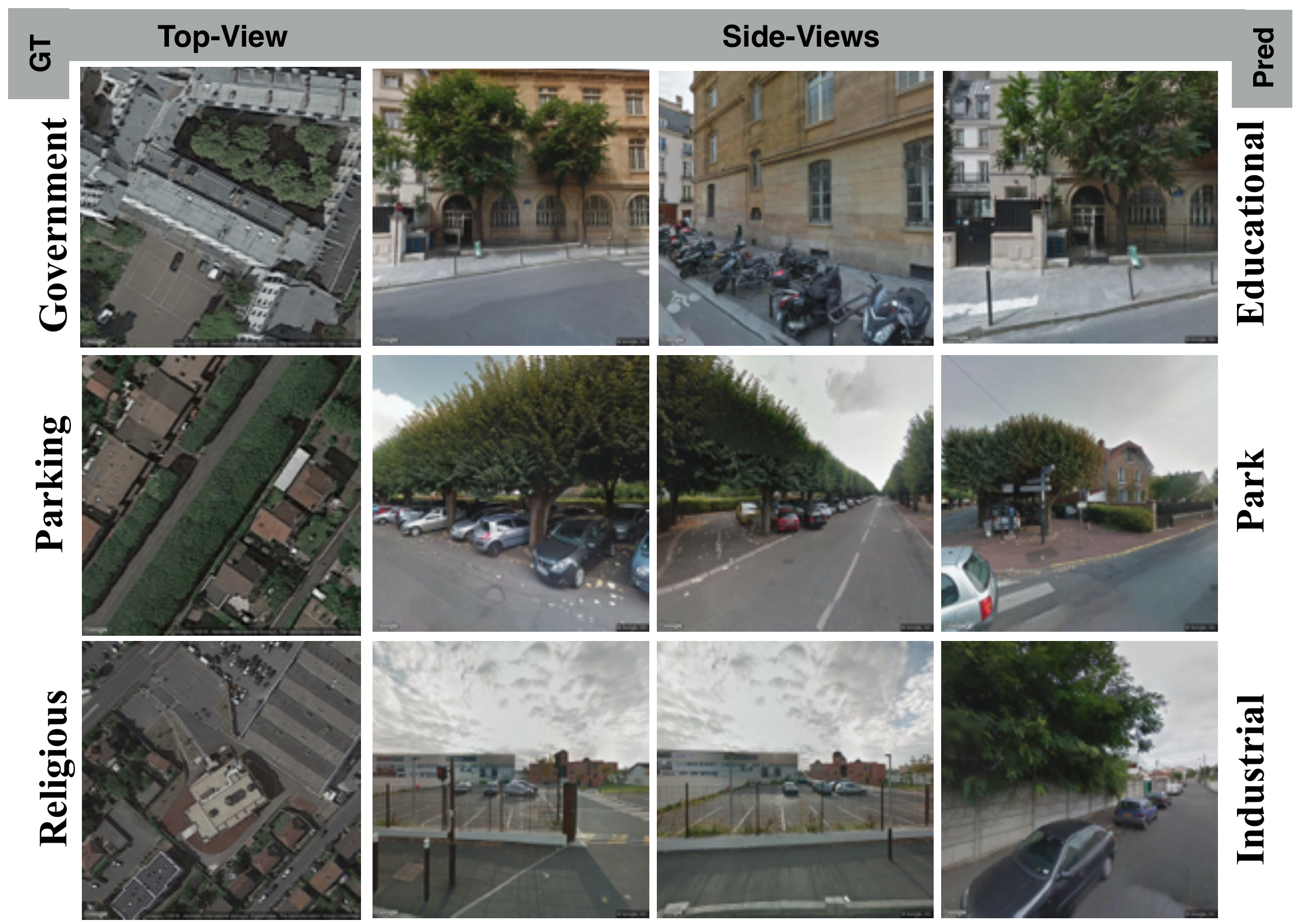}
\caption{\label{fig:negative_results}Examples of wrongly classified urban-objects by the proposed multimodal CNN model. For each row, the ground truth class is mentioned on the left hand side, while the predicted class is shown on the right hand side. Regarding the images, the first column shows overhead imagery, while the other three come from the ground-based collection.}
\end{figure}

Some misclassifications are shown in Figure~\ref{fig:negative_results}. For example, the model predicts class \emph{educational} for the ``government" urban-object in the first row (Figure~\ref{fig:negative_results}). {This} most probably {emerges from} the presence of information similar to that of an educational place in the ground views, such as the presence of objects like open spaces and benches in front or metallic fences enclosing the building. The second row of Figure~\ref{fig:negative_results} shows a parking area that has been predicted as a park, most likely due to the many trees visible in both the top and the ground views. In the third row of Figure~\ref{fig:negative_results}, the urban-object with class \emph{religious} was predicted as an \emph{industrial} facility, possibly due to the large parking area with cars as seen in both the top and the side views, while the church far in the distance is vaguely visible. \shiva{}{Wrong label predictions are sometimes  observed because of the low quality of the downloaded GSV pictures. We found two issues about the downloading of GSV pictures for OSM polygons: i) in some cases the OSM polygons do not match with the actual boundaries of the urban-objects and ii) the distance-based heuristic used to download GSV pictures is sometimes inaccurate and
leads to the download of pictures of other nearby urban-objects. These issues are also discussed in}~\cite{srivastava2018_ijgis}.

\setlength{\tabcolsep}{2pt}
\begin{table}[t]
\begin{center}
\caption{\shiva{}{Accuracy scores for our proposed Multimodal CNN and VIS-CNN using different base models (ResNet50 and AlexNet instead of VGG16) for {\^I}le-de-France. OH: overhead imagery, GSV: Google Street View ground based images. OA: overall accuracy; AA: average accuracy). }}
\label{table:accuracy_scores_other_cnns}
\begin{tabular}{l|c|c|cc}
\hline\noalign{\smallskip}
&\multicolumn{2}{c|}{Data source(s)}& \multicolumn{2}{c}{Metric} \\
Model Name & Train & Test  &  OA & AA\\
\noalign{\smallskip}
\hline
\noalign{\smallskip}
AlexNet~\cite{krizhevsky2012imagenet}& OH & OH & 63.42 $\pm$ 1.35 & 57.45 $\pm$ 1.44 \\
ResNet50~\cite{He_2016}& OH & OH & 67.53 $\pm$ 1.07 & 64.18 $\pm$1.54 \\
\hline
VIS-CNN with Avg, AlexNet& GSV & GSV & 57.13 $\pm$ 1.18 & 54.10 $\pm$ 0.82 \\
VIS-CNN with Avg, ResNet50& GSV & GSV & 54.60 $\pm$ 2.62  & 54.95 $\pm$ 3.81  \\
\hline
Multimodal CNN, AlexNet& OH, GSV & OH, GSV & 69.21 $\pm$ 0.64 & 66.44 $\pm$ 0.92 \\
Multimodal CNN, ResNet50& OH, GSV & OH, GSV & 68.96 $\pm$ 0.89 & 67.25 $\pm$ 1.44\\
\hline
\end{tabular}
\end{center}
\end{table}

\begin{figure}[t]
\centering
\includegraphics[width=0.9\textwidth]
{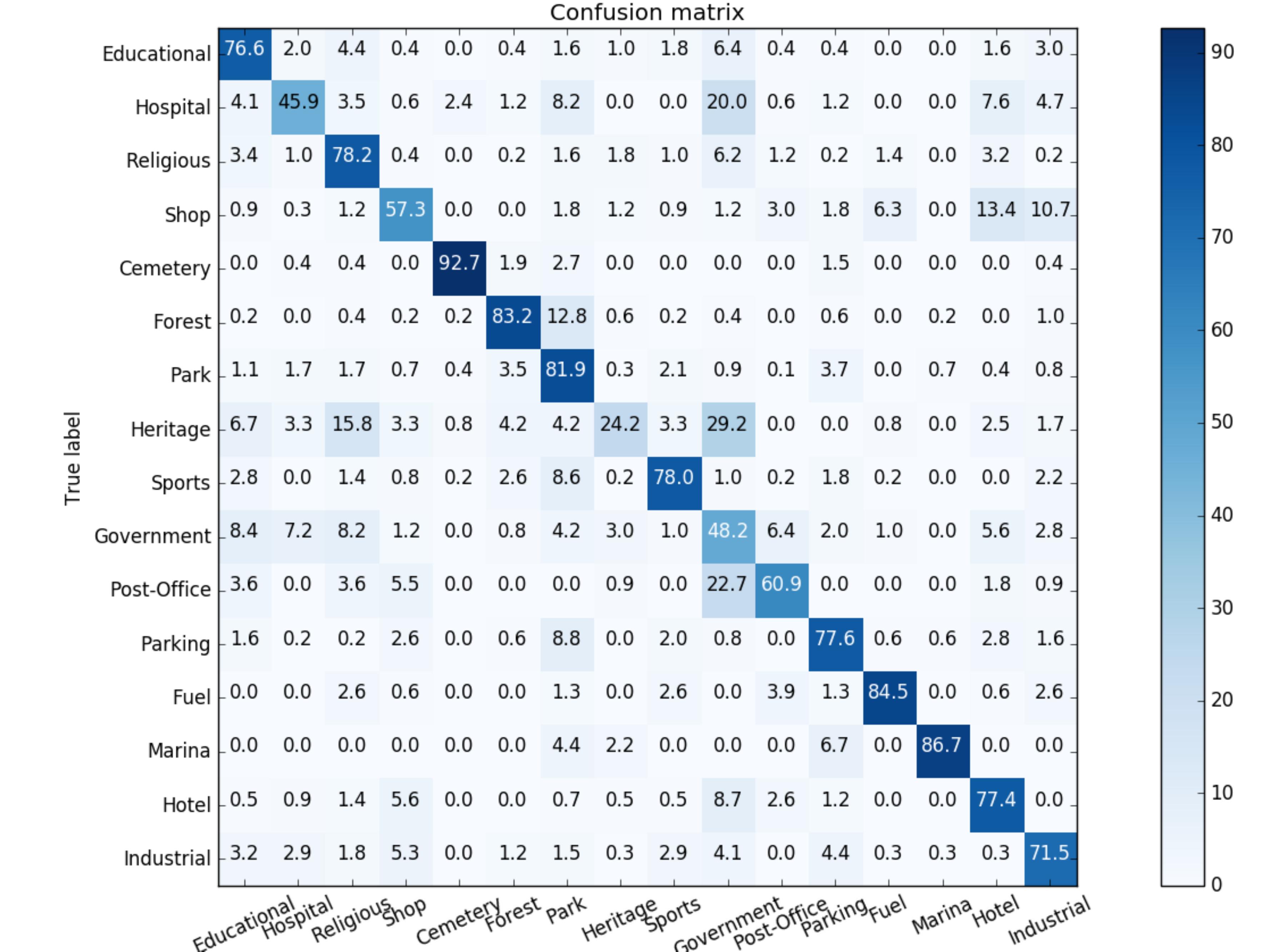}
\caption{\label{fig:conf_matrix_all_splits}\shiva{}{Confusion Matrix (values in \%) for the prediction results in the {\^I}le-de-France dataset. We average the confusion matrices of all the evaluated five test splits. To obtain the percentage values we divided the number of samples of each cell by the total number of samples of its corresponding row and then multiplied by 100.}}
\end{figure}

\shiva{}{In order to show in more detail the accuracy of the model for each class, in Figure~\ref{fig:conf_matrix_all_splits} we present the confusion matrix generated by averaging the test accuracy of the Multimodal CNN method (with VGG16 as base CNN model) for Ile-de-France dataset. We can see that classes like ``Hospital", ``Heritage", and ``Post-Office" are often wrongly predicted as class ``Government". We can also observe that the urban-objects of ``Forest" are sometimes classified as ``Park" and urban-objects of "Shop" are occasionally misclassified as ``Hotel". }

\subsection{\shiva{}{Generalisability of the model in a new city}}

\shiva{}{We have used the data from the city of Nantes to evaluate the generalisation ability of our model. In Table~\ref{tab:nantes_vgg16} we present the OA and AA scores of the proposed Multimodal CNN model and the two unimodal models, trained with Ile-de-france data. Overall, the model provides results in the ballpark of those observed for {\^I}le-de-France. AA scores are generally lower, mostly because the `Marina' class omitted for this dataset was very accurate in the {\^I}le-de-France case (average of 86\% Producer accuracy, see Figure~\ref{fig:conf_matrix_all_splits}). Comparing the methods in the Nantes case, the proposed Multimodal CNN is 5\% more accurate in OA and 10\% in AA with respect to the model that uses only overhead imagery. It also improves the accuracy of VIS-CNN by more than 16\% in OA and 11\% in AA, once again confirming the observations made in the first dataset. Note that we ran inference on the Nantes urban-objects directly, without finetuning any further the models.}

\begin{table}[t]
\centering
\caption{\shiva{}{Accuracy scores of the proposed Multimodal CNN model and two unimodal CNN models for the city of Nantes} \label{tab:nantes_vgg16}}
\begin{tabular}{p{3.5cm}|c|c|c}
 \hline
 Base Model & Data Source (Test) & OA & AA\\
 \hline
 VGG16 & OH & 70.94 $\pm$ 0.44& 53.9 $\pm$ 1.13\\
 VIS-CNN with Avg & GSV  & 58.54 $\pm$ 0.72 & 52.11 $\pm$ 0.80\\
  Multimodal CNN & OH, GSV & \bf{75.07} $\pm$ 1.10& \bf{62.91} $\pm$ 0.75\\
 \hline
\end{tabular}

\end{table}

\subsection{Missing modality retrieval}\label{ssec:ccares}
\newcommand{\sss}{0.2}

In this section, we test the ability of our model to predict landuse when the GSV pictures are missing. To do so, we use the CCA-based system presented in Section~\ref{ssec:cca}. 

\begin{figure*}
\begin{center}
\begin{tabular}{cccc} 
  \includegraphics[width=\sss\columnwidth]{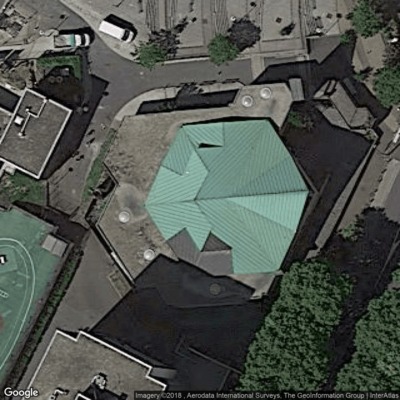} & 
  \includegraphics[width=\sss\columnwidth]{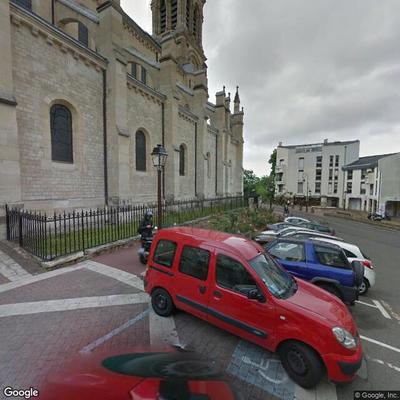} & 
  \includegraphics[width=\sss\columnwidth]{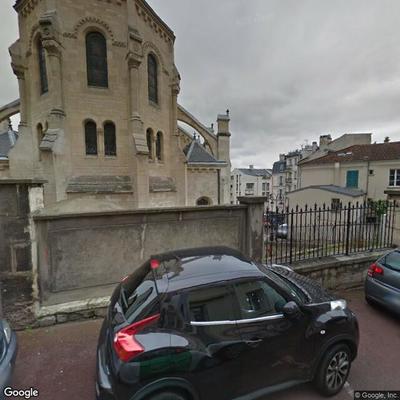} & 
  \includegraphics[width=\sss\columnwidth]{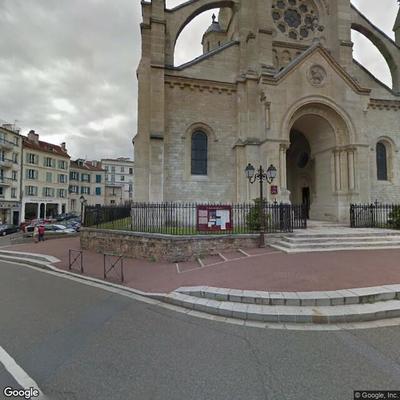} \\
  (a) Overhead: Religious  & \multicolumn{3}{c}{GSV : Religious}\\ 
  \includegraphics[width=\sss\columnwidth]{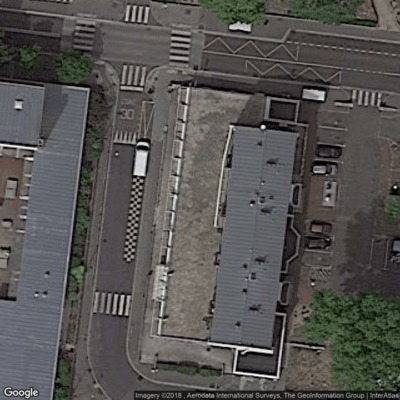} & 
  \includegraphics[width=\sss\columnwidth]{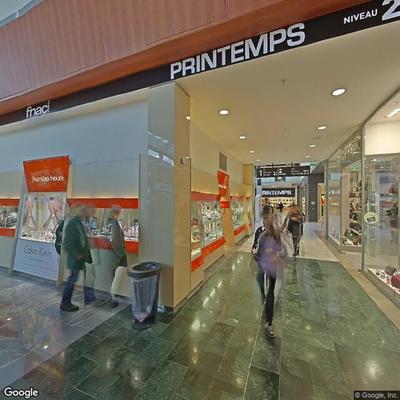} & 
  \includegraphics[width=\sss\columnwidth]{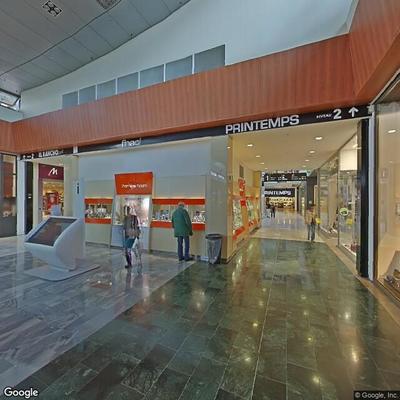} & 
  \includegraphics[width=\sss\columnwidth]{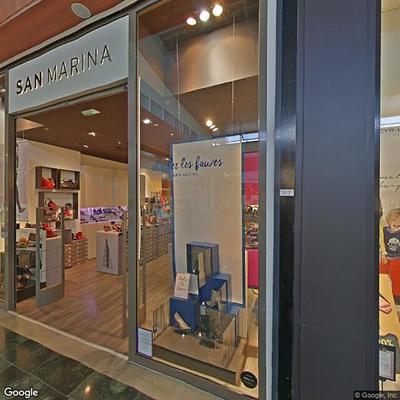} \\
  (b) Overhead Shop  & \multicolumn{3}{c}{GSV : Shop}\\ 
    \includegraphics[width=\sss\columnwidth]{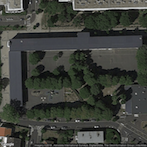} & 
  \includegraphics[width=\sss\columnwidth]{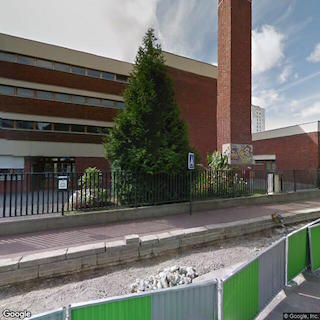} & 
  \includegraphics[width=\sss\columnwidth]{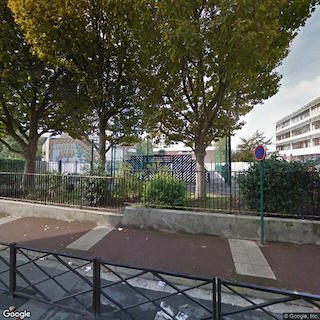} & 
  \includegraphics[width=\sss\columnwidth]{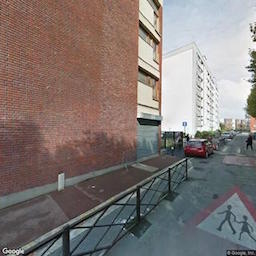} \\
   (c) Overhead: Education  & \multicolumn{3}{c}{GSV : Education}\\
  \includegraphics[width=\sss\columnwidth]{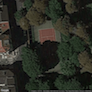} & 
  \includegraphics[width=\sss\columnwidth]{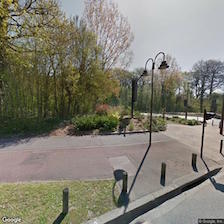} & 
  \includegraphics[width=\sss\columnwidth]{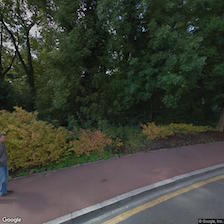} & 
  \includegraphics[width=\sss\columnwidth]{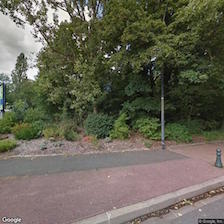} \\
  (d) Overhead: Sports  & \multicolumn{3}{c}{GSV : Forest}\\
  \includegraphics[width=\sss\columnwidth]{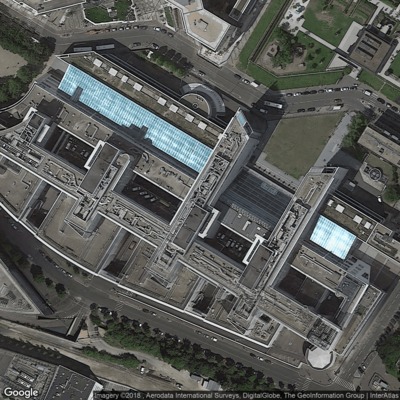} & 
  \includegraphics[width=\sss\columnwidth]{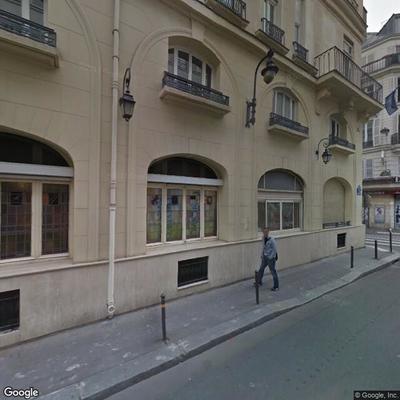} & 
  \includegraphics[width=\sss\columnwidth]{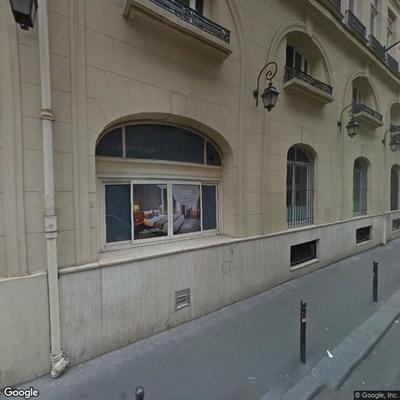} & 
  \includegraphics[width=\sss\columnwidth]{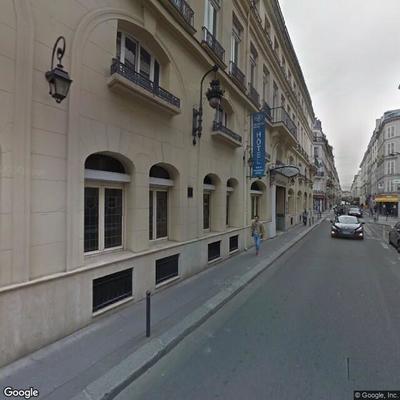} \\
  (e) Overhead: Hospital  & \multicolumn{3}{c}{GSV : Hotel}\\
\end{tabular} 
\end{center}
\caption{Examples of retrieved GSV pictures for a given query overhead imagery. The overhead imagery is shown in the first column, corresponding sets of retrieved GSV pictures are shown in columns 2 to 4. (a), (b) and (c) are retrieval results of the correct class, while (d) and (e) are retrievals of an incorrect class.\label{fig:GSV_from_top_ranked_OSM_for_OH}}
\end{figure*}

\subsubsection{Numerical performance}
The overall results are reported in the last row of Table~\ref{table:accuracy_scores}, which shows the accuracy obtained by retrieving the missing GSV pictures for an urban-object that just have an overhead imagery and then performing the label prediction using the proposed multimodal model. We can observe that the accuracy obtained by this method is higher by more than 4\% in OA compared to the model that just uses overhead imagery (Section~\ref{ssec:overheadCNN}). 

Figure~\ref{fig:GSV_from_top_ranked_OSM_for_OH} shows examples of retrieved GSV pictures (corresponding to urban-objects with the highest similarity scores) for five different overhead images. The first three rows show positive examples, with retrieved GSV pictures belonging to the same class as the queried overhead imagery. In these three examples, the retrieved ground-based pictures have discriminative visual features that can help to predict the correct labels when using the multimodal model, even though they come from another urban-object. The fourth and fifth row present negative retrieval examples, were the retrieved GSV pictures belong to a different class compared to the queried overhead imagery. Note that the overhead image in the fourth row belongs to class ``sports" as it contains a tennis court. However, since it is occluded by trees, the closest GSV pictures that were retrieved belonged to the class ``forest".

\begin{figure}[!t]
\centering
\includegraphics[width = \linewidth]{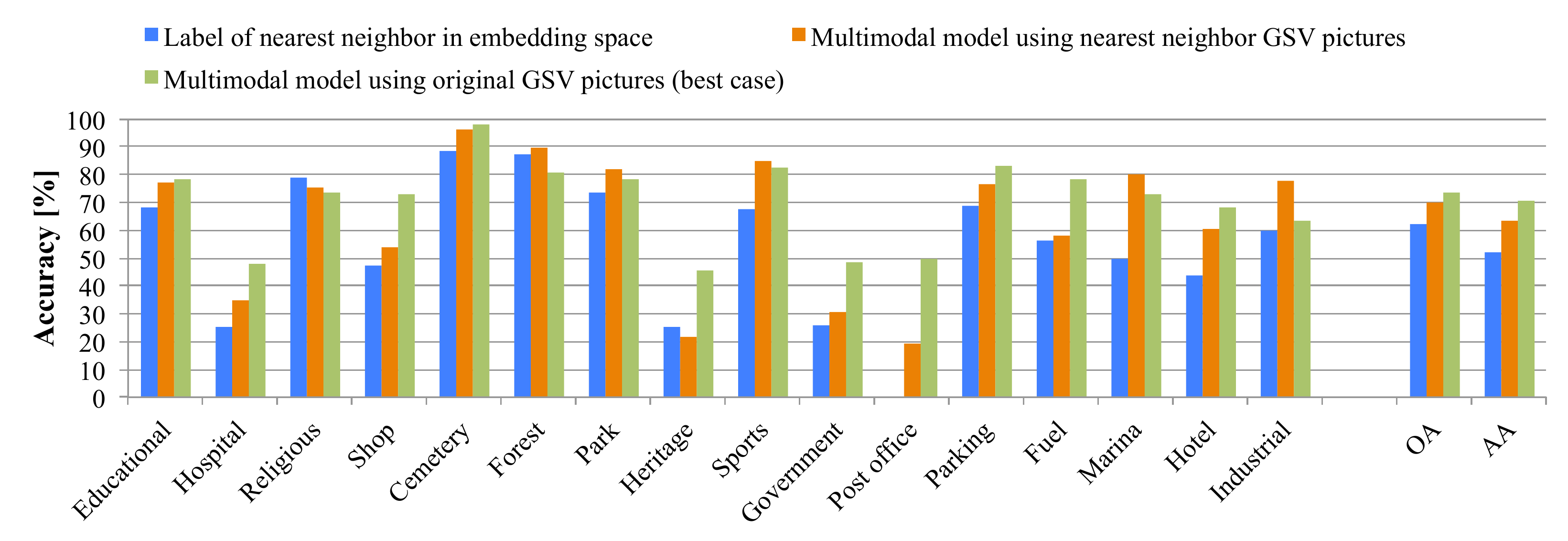}
\caption{Numerical results of the experiment considering samples without GSV pictures. In blue: accuracies of the labels of the nearest neighbors in the embedding space; in orange: results of the multimodal model using the retrieved GSV pictures of the nearest neighbor in the GSV stream; in green: results of the full model, using the real GSV pictures for the test urban-object. All per class scores are producer's accuracies (\% that a class is predicted correctly with respect to the total of the ground truth labels of that class) \shiva{}{for one single run with the same seed}.\label{fig:perclass}}
\end{figure}

Figure \ref{fig:perclass} shows the classification results per class in terms of producer's accuracy for one run of the algorithm. One can appreciate the accuracy of the direct retrieval of the nearest neighbors labels (blue bars), which is around $70\%$ for seven out of the sixteen classes.  Poor results are obtained for classes `Hospital', `Heritage' and `Post office'. These classes correspond to those with less examples in the training set. Using the GSV pictures of the retrieved training objects together with the true overhead images in the multimodal model  (orange bars, corresponding to our proposition) strongly improves the results and almost closes the gap with the full multimodal model (green bars). The latter is an upper bound on performance, since it uses the real GSV pictures. The classes for which the accuracy of the full model is not matched correspond to those with low number of samples, which already had a poor retrieval accuracy in the embedding space.

\begin{figure}[!t]
\centering
\includegraphics[width = \linewidth]{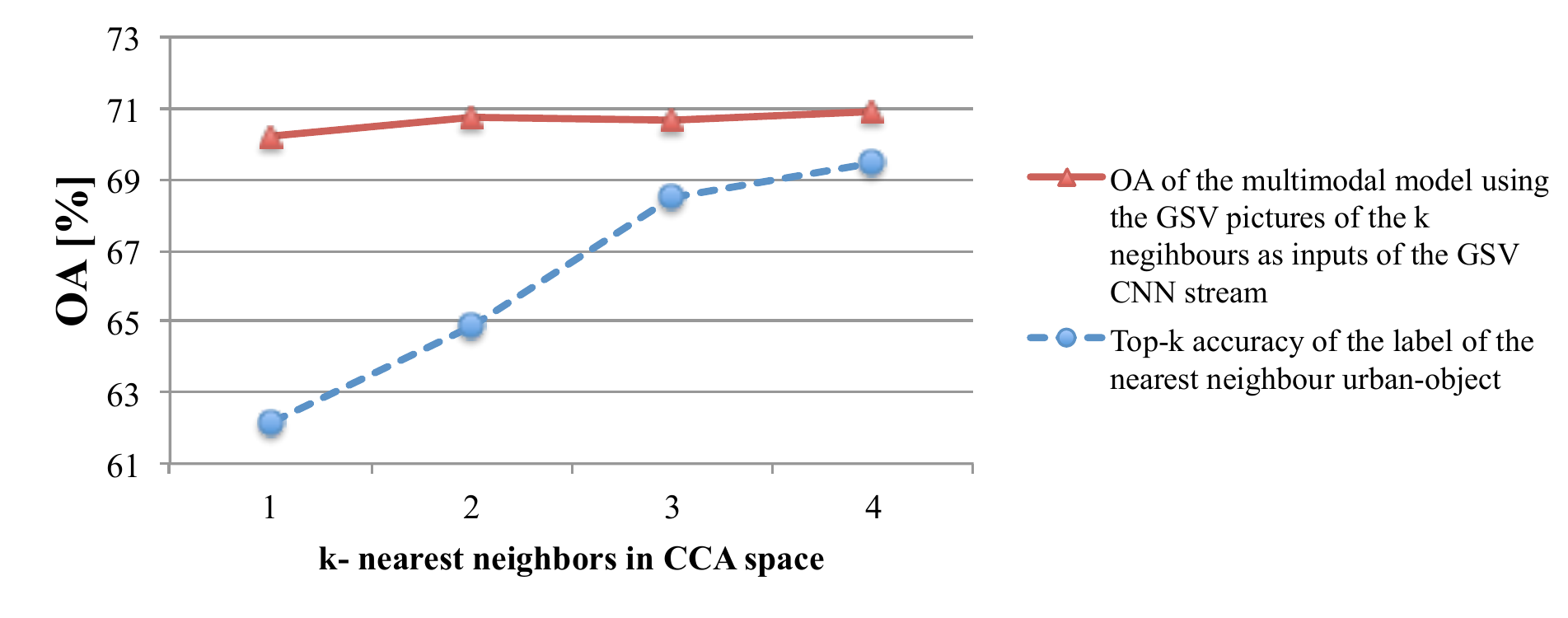}
\caption{Blue: top-$k$ accuracy of retrieval in the CCA embedding space: it corresponds to the number of times training urban-objects of the correct class are among the first $k$ nearest neighbors. Red: overall accuracy of the multimodal CNN, when predicting using both the original overhead image and the GSV pictures of the $k$-nearest neighbors retrieved.}\label{fig:coherence}
\end{figure}

\subsubsection{Label coherence in the embedding space}
To follow up this last observation, we analyze the label coherence in the embedding space, i.e. we want to verify that the urban-objects without GSV pictures are projected close to other urban-objects of the correct class. The blue curve in Figure~\ref{fig:coherence} illustrates the trend for an increasing number of nearest neighbors (i.e. a $top-k$ accuracy). After projection, the test urban-object is mapped close to a sample of the correct class $62\%$ of the times, but this percentage increases when considering more neighbors in the embedding space (up to $69\%$ of the test samples are mapped close to at least one training sample of the correct class): this shows that the CCA space is coherent in terms of labels and that the retrieval can be successful. However, such increase in top-$k$ accuracy has surprisingly little influence of the performance of the final multimodal model (red solid curve in Figure~\ref{fig:coherence}): even when using GSV pictures of the four nearest neighbors in the CCA space, the increase in performance is of $1\%$ only. We believe this modest increase in performance is due to the fact that, even though at least one training urban-object retrieved is of the right class, at most $k-1$ others will be of an incorrect class, which might confuse the GSV stream and impede larger improvements. To support this hypothesis, we evaluated the average number of nearest neighbors of the correct class: $0.65$ for $k=1$,  $1.22$ for $k=2$, $1.85$ for $k=3$ and $2.46$ for $k=4$. Therefore, for smaller values of $k$, the GSV stream will receive pictures from objects of the right class approximately 60\% of the times, which allows it to provide a robust response leveraging the discriminative information in the overhead view.

\subsubsection{Sensitivity to the parameters of the CCA model}
\begin{figure}[!t]
\begin{center}
\begin{tabular}{cccc} 
  \rotatebox{90}{} &
  \includegraphics[width=0.33\columnwidth]{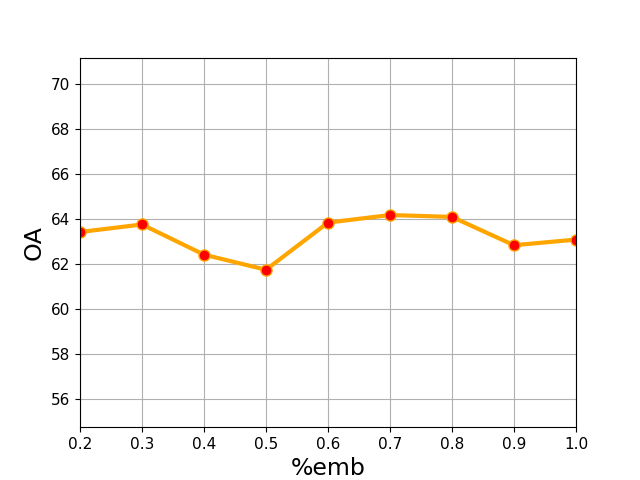} & 
  \includegraphics[width=0.33\columnwidth]{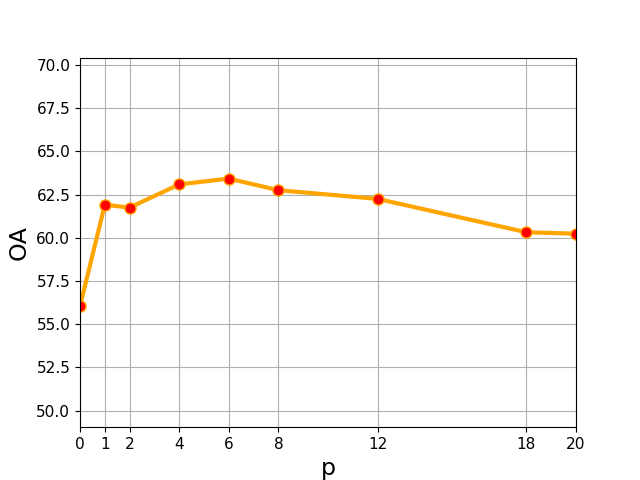} & 
  \includegraphics[width=0.33\columnwidth]{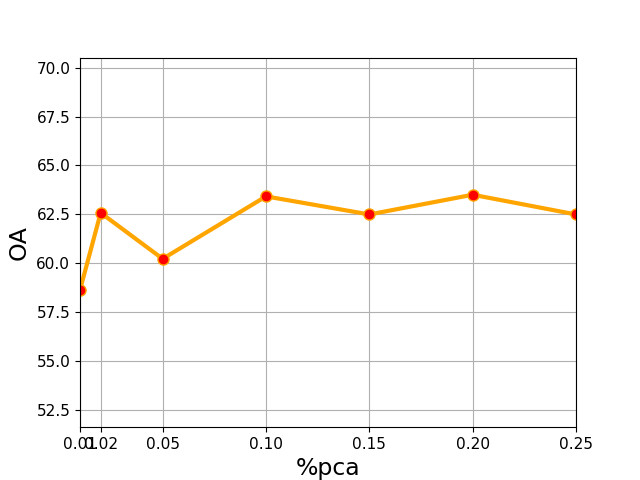} \\
  &(a) $\%d_{emb}$ & (b) $p$ & (c) $\%pca$\\
\end{tabular} 
\end{center}
\caption{Sensitivity study for the hyperparameters of the CCA three-view embedding space. The accuracies are computed using the label of the closest neighbor among the training GSV objects as prediction. When varying one parameter, the two others are fixed to the following values: $\%d_{emb} = 0.2$, $p = 6$ and $\%pca = 0.1$.
\label{fig:ablation_study_noLtrain}}
\end{figure}

Finally, we provide an analysis of the sensitivity of the CCA model to its free parameters. For the results in fourth row of Table~\ref{table:accuracy_scores}, we empirically selected the parameter values of the proposed method: \%pca $= 0.1$,  \%$d_{emb} = 0.2 $ and  $p = 6$. Figure~\ref{fig:ablation_study_noLtrain} shows the overall retrieval accuracy when fixing two of the three parameter values and varying the values of the third. These accuracies were computed by projecting the overhead imagery features of the test set into the embedding space and using the label of the nearest urban-object of the training set for prediction. We observed that the proposed system behaves in a stable manner when varying the hyperparameters.

\section{Conclusions and Outlook}\label{sec:conclusions}

In this work, we presented a multimodal model for landuse classification that uses pictures from top and ground views with annotations from OpenStreetMap. The proposed model learns end-to-end both the feature extraction from single modalities and their fusion. We evaluated our proposed method in the region of {\^I}le-de-France, France and found that, for many classes, the complementary visual information contained in either modality improved the accuracy of the model by a large margin. Our proposed multimodal CNN model can also predict landuse labels when ground-based pictures are not available for an urban-object by searching for the most plausible set of GSV pictures in the training set. 

Using widely available data repositories for images (Google Street View and Google Maps) and public participatory vector annotations (OpenStreetMap) gives an edge to our model, as it is scalable to several other cities. The accuracies could be further improved by having a better quality dataset. This could be achieved by sourcing better quality labels (e.g., labels from other sources like Google Places) and/or refining heuristics for downloading the GSV pictures (e.g., collecting pictures that are looking at the urban-objects' facade more accurately). For future work, we plan to explore the image information available at multiple scales as an input for our proposed model, as well as integrating fine-grained object detection in the ground images (e.g. objects like ambulances) as extra information cues.

\section*{Acknowledgment}
The authors would like to thank Google and OSM for the access to pictures and objects' footprints respectively through their APIs. This work has been supported by the Swiss National Science Foundation (grant PZ00P2-136827 (SS,DT, \url{http://p3.snf.ch/project-136827}). JEVM acknowledges FAPESP (grant 2016/14760-5, 2017/10086-0) for support.

\section*{References}
\bibliography{main}

\end{document}